\documentclass[letterpaper]{article} % DO NOT CHANGE THIS
\usepackage{aaai24}  % DO NOT CHANGE THIS
\usepackage{times}  % DO NOT CHANGE THIS
\usepackage{helvet}  % DO NOT CHANGE THIS
\usepackage{courier}  % DO NOT CHANGE THIS
\usepackage[hyphens]{url}  % DO NOT CHANGE THIS
\usepackage{graphicx} % DO NOT CHANGE THIS
\usepackage{natbib}  % DO NOT CHANGE THIS AND DO NOT ADD ANY OPTIONS TO IT
\usepackage{caption} % DO NOT CHANGE THIS AND DO NOT ADD ANY OPTIONS TO IT
\usepackage{algorithm}
\usepackage{algorithmic}
\usepackage[dvipsnames]{xcolor}

\usepackage{makecell}
\usepackage{booktabs}
\usepackage{arydshln}
\usepackage{amsmath}
\usepackage{newtxtext,newtxmath}
\usepackage{xcolor}
\usepackage{lipsum}  
\usepackage{subcaption}
\usepackage{multirow}
\usepackage{hhline}

\usepackage{newfloat}
\usepackage{listings}
\DeclareCaptionStyle{ruled}{labelfont=normalfont,labelsep=colon,strut=off} % DO NOT CHANGE THIS
\floatstyle{ruled}
\newfloat{listing}{tb}{lst}{}
\floatname{listing}{Listing}
\usepackage{multirow}
\title{Improving Transferability for Cross-domain Trajectory Prediction \\ via Neural Stochastic Differential Equation}
\author{
    Daehee Park, Jaewoo Jeong, and Kuk-Jin Yoon
}
\affiliations{
    Visual Intelligence Lab., KAIST, Korea\\
    \{bag2824, jeong207, kjyoon\}@kaist.ac.kr
}

\begin{document}

\maketitle

\begin{abstract}
Multi-agent trajectory prediction is crucial for various practical applications, spurring the construction of many large-scale trajectory datasets, including vehicles and pedestrians. 
However, discrepancies exist among datasets due to external factors and data acquisition strategies. External factors include geographical differences and driving styles, while data acquisition strategies include data acquisition rate, history/prediction length, and detector/tracker error. 
Consequently, the proficient performance of models trained on large-scale datasets has limited transferability on other small-size datasets, bounding the utilization of existing large-scale datasets.
To address this limitation, we propose a method based on continuous and stochastic representations of Neural Stochastic Differential Equations (NSDE) for alleviating discrepancies due to data acquisition strategy. 
We utilize the benefits of continuous representation for handling arbitrary time steps and the use of stochastic representation for handling detector/tracker errors. 
Additionally, we propose a dataset-specific diffusion network and its training framework to handle dataset-specific detection/tracking errors. 
The effectiveness of our method is validated against state-of-the-art trajectory prediction models on the popular benchmark datasets: nuScenes, Argoverse, Lyft, INTERACTION, and Waymo Open Motion Dataset (WOMD). 
Improvement in performance gain on various source and target dataset configurations shows the generalized competence of our approach in addressing cross-dataset discrepancies.
\let\thefootnote\relax\footnotetext{The code is available at \url{https://github.com/daeheepark/TrajSDE}.}
% \blfootnote{The code is available at https://github.com/daeheepark/TrajSDE.}
% The code is available at \url{https://github.com/daeheepark/TrajSDE}
% \blfootnote[right]{Right bottom footnote text.}

\end{abstract}

\section{Introduction}
% Multi-agent의 미래 경로를 예측하는 것은 자율주행 시스템에서 안전한 planning을 위해 매우 중요한 분야이다.
Trajectory prediction stands as one of the most crucial challenges to corroborate the safety of autonomous driving systems. 
% Trajectory prediction은 detection/tracking 에 의해 찾아진 tracklet history로부터 미래 수초간의 미래 경로를 예측하는 것이 목적이다.
Its objective of predicting future trajectories allows autonomous agents to respond optimally to actively changing environments. 
% 자율주행차는 미래의 위험을 사전에 인지하고 대비하는 것이 중요하므로, 주변 차량의 미래 경로를 정확하게 예측해내는 것이 중요하다.
% As individual agents need to perceive and respond to potential hazards proactively, accurate prediction of the future paths of surrounding agents becomes paramount.
% 따라서 최근에 데이터 driven 기반으로 이 문제를 해결하려는 시도가 굉장히 많이 있었다.
As a response, a number of large-scale trajectory datasets such as nuScenes, Argoverse, WOMD, Lyft, INTERACTION, and TrajNet++ have been established~\cite{caesar2020nuscenes, chang2019argoverse, interactiondataset,houston2021one, ettinger2021large, kothari2021human} to pursue a data-driven approach towards constructing a reliable motion forecasting system~\cite{li2021grin, tang2021collaborative, Bae2023ASO, Wu2023MultiStreamRL, Ge2023CausalIF, shi2022social, liang2021temporal}.

% Consequently, recent efforts have been directed towards data-driven approaches to tackle this problem
% 이러한 목적을 위해 nuScenes, Argoverse2, Waymo Open Motion Dataset (WOMD) 과 같은 여러 large-scale trajectory dataset이 구축되었다.

% Data driven model 의 잘 알려진 문제는 학습한 데이터셋과 다른 distribution을 가진 환경에서는 잘 동작하지 않는다는 것이다.
% 따라서 누군가가 자신만의 환경에서 trajectory prediction system을 만드려고 할 때, 기존에 구축된 large-scale labeled dataset을 제대로 활용하지 못한다는 단점이 있다.
% 이에 따라 최근에는 trajectory prediction task에서 cross-domain learning 을 시도한 논문이 다수 나왔다.
% 그들은 federated learning을 사용하여 multi-source dataset에서 학습하여 모델의 generality를 높이거나, 혹은 domain adaptaion 방법을 제안하였다.
% ~~는 source dataset과 target dataset의 feature 분포 사이의 거리를 가깝게 하는 일반적인 da 방식을 하용하였다.
% ~~는 주행경로가 취득된 공간에 대한 generality를 높이는 방법을 제안하였다.
% 하지만 trajectory prediction dataset에는 이런 discrepancy 외에도 다른 특수한 discrepancy가 존재한다.

\begin{figure}[!t]
\centering
\begin{subfigure}{0.325\columnwidth}
    \includegraphics[width=\linewidth]{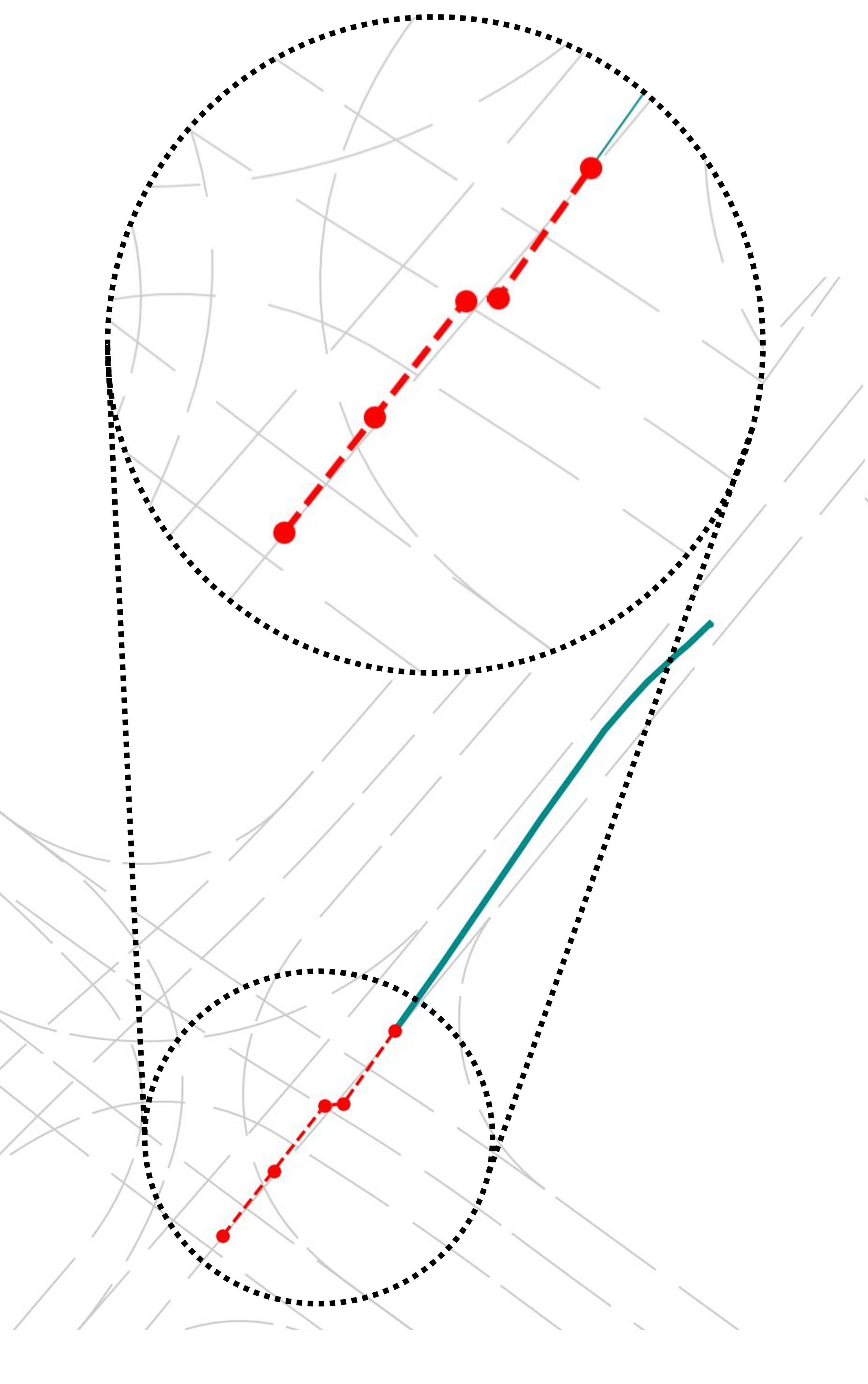}
    \caption{nuScenes}
    \label{fig:first}
\end{subfigure}
\hfill
\begin{subfigure}{0.325\columnwidth}
    \includegraphics[width=\linewidth]{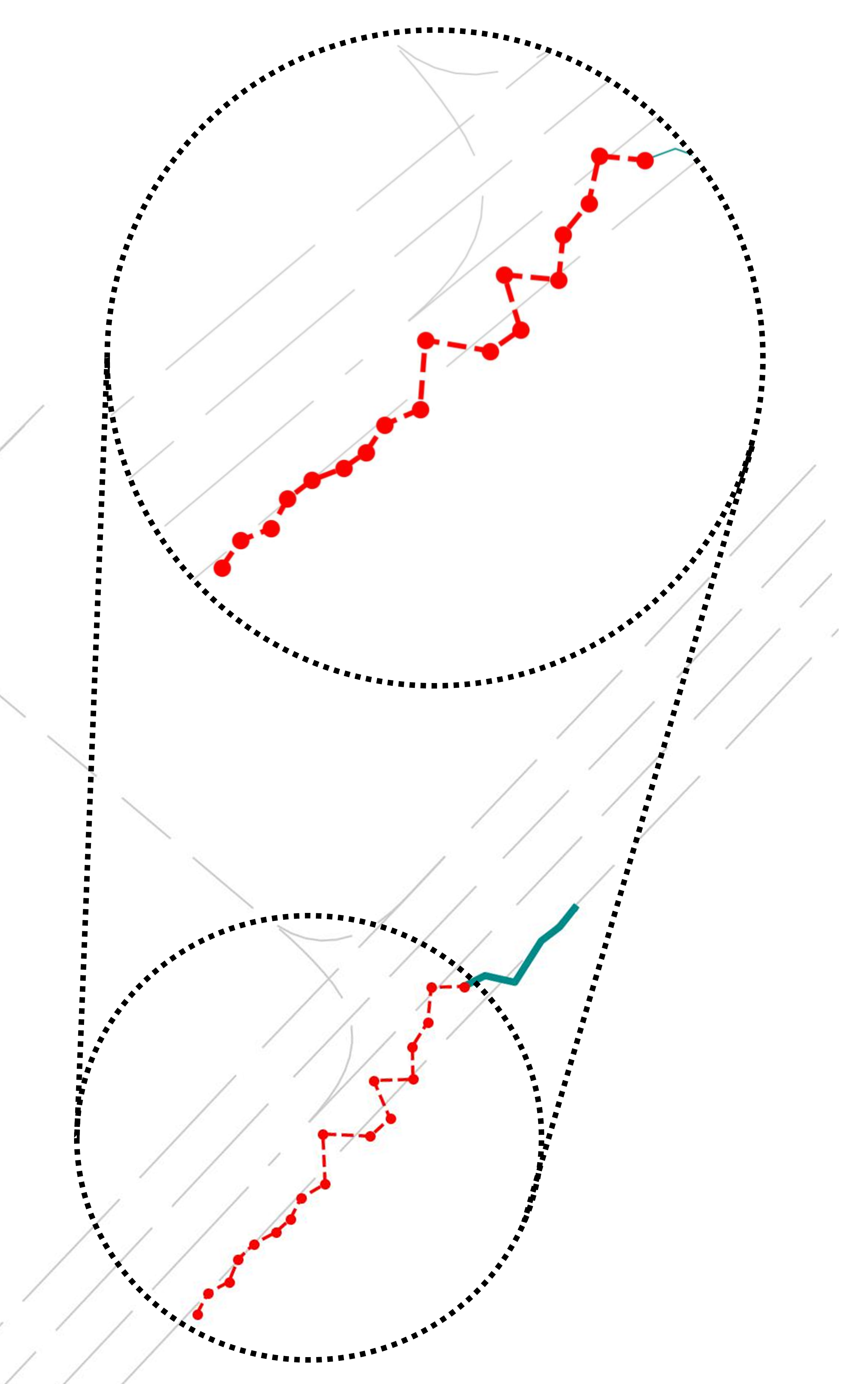}
    \caption{Argoverse}
    \label{fig:second}
\end{subfigure}
\hfill
\begin{subfigure}{0.325\columnwidth}
    \includegraphics[width=\linewidth]{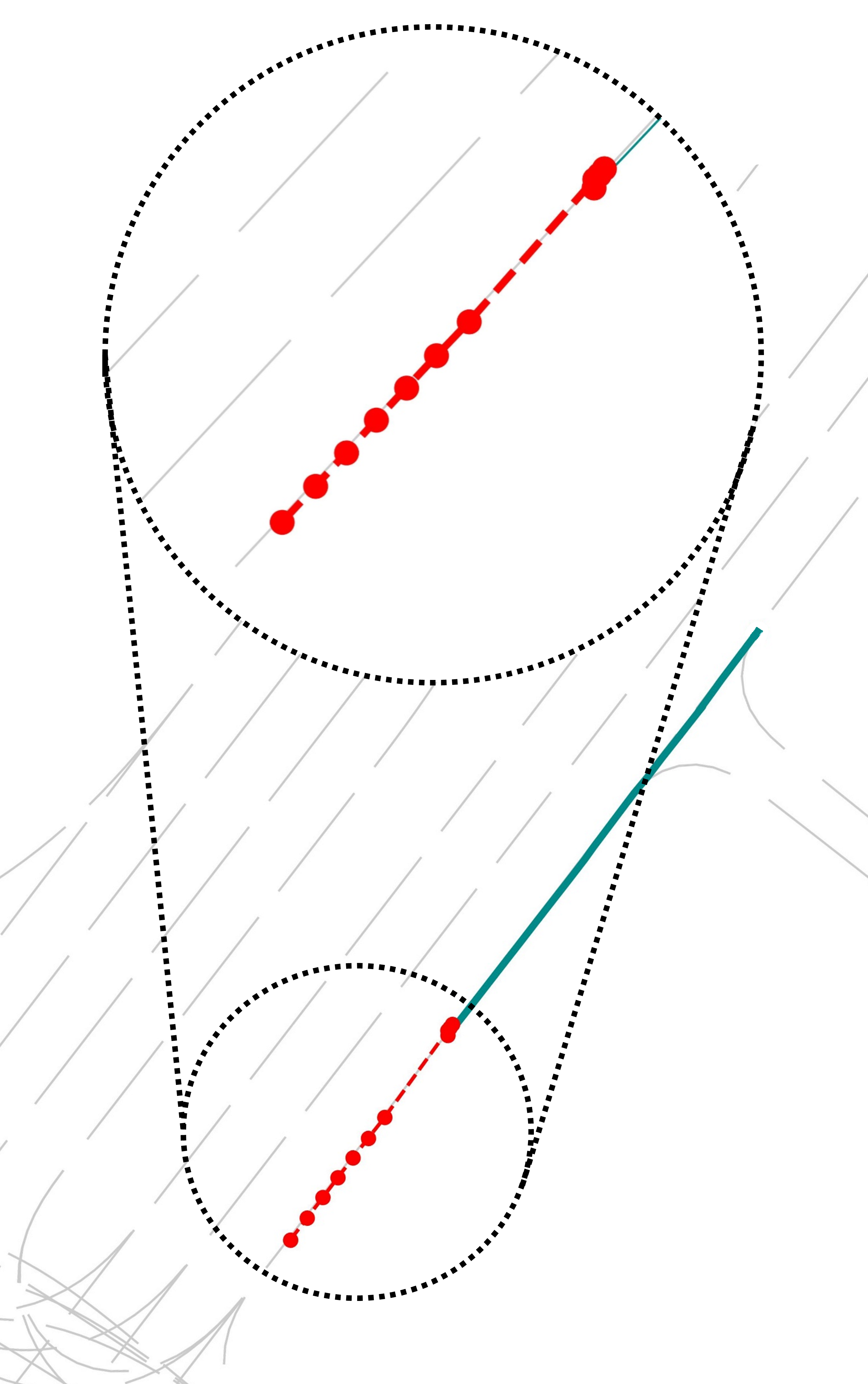}
    \caption{WOMD}
    \label{fig:third}
\end{subfigure}        
\caption{Unique uncertainty manifested across each trajectory prediction dataset due to discrepancy in data acquisition strategy. The \textcolor{red}{red} dotted line and \textcolor{JungleGreen}{darkgreen} solid line represent past and future trajectories. The two main sources of discrepancy are time step configuration difference and tracklet noise. Tracklet errors are uniquely shown as lateral position error in nuScenes, ID switch in Argoverse, and longitudinal position error in WOMD, all of which our framework handles in a dataset-wise exclusive manner.}
\label{fig:intro_datadiff}
\end{figure}
\begin{table}[!t]
\small
\begin{tabular}{lcccc}
\toprule
                           & \small\makecell{History \\ (s)} & \small\makecell{Prediction \\ horizon (s)} & \small \makecell{Frequency \\ (Hz)} & \small \makecell{Training \\ size} \\ \midrule
nuScenes    & 2                                & 6                                           & 2                                   & 30k           \\ 
Argoverse   & 2                                & 3                                           & 10                                  & 200k          \\ 
WOMD        & 1                                & 8                                           & 10                                  & 500k          \\
TrajNet++     & 3.2                               & 4.8                                       & 2.5                                   & 250k              \\
\bottomrule
\end{tabular}
\caption{Various temporal configurations during data acquisition across trajectory datasets.
These discrepancies - 1) past/future time length and 2) data acquisition rate - severely limit cross-dataset transferability.}
% \vspace{-10pt}
\label{tab:intro_timestep_diff}
\end{table}

One well-known issue with data-driven models is their limited performance when discrepancies in data distributions are manifested between training and test data. 
Therefore, to construct a trajectory prediction system on a specific environment, the optimal way is to collect data from that environment.
However, recent models require abundant data for optimal performance, which require a cumbersome process of acquiring such an amount of data.
In that sense, adequate utilization of existing large-scale datasets grants an advantage in circumventing this hurdle.
% As a result, when individuals aim to create their own trajectory prediction system for a specific environment, they are unable to fully harness the abundant data of existing large-scale labeled datasets.
% Indeed, discrepancies in data distributions limit the potential of transferring the knowledge trained on a larger dataset to the model on a smaller dataset.
Recent approaches have attempted to overcome such challenge by proposing domain adaptation~\cite{9880042, wang2022transferable} or increasing model generalizability via multi-source dataset training~\cite{Wang_2022_CVPR}.
% using cross-domain or multi-source learning~\cite{9880042, wang2022transferable, Wang_2022_CVPR} by increasing generalizability.
% Furthermore, some researchers have suggested techniques to increase generality concerning the space where driving paths are acquired~\cite{Ye2023ImprovingTG}. 
Compared to these efforts in handling domain gaps, the dataset-specific discrepancies caused by disparity between each data acquisition strategy have been excluded from being considered as a domain gap and have been less visited. 
Our work shows that adequate handling of these dataset-specific discrepancies unlocks a collective potential from cross-dataset motion patterns.
% we achieve superior performance in a specific target dataset of interest based on motion patterns from other datasets.

% 첫번째로, 데이터셋마다 주어지는 과거 경로의 시간 길이와 예측해야하는 미래 경로의 시간 길이가 상이하다.
% 이에 따라 각 time configuration에 따른 주행 경로 feature의 manifold가 feature space 상에서 달라질 수 있다.
% 예를 들어 1초 관측 데이터로부터 8초 미래를 예측하는 Waymo 데이터셋에서 학습된 모델은 과거 1초 경로로부터 8초 미래 경로를 나타내는 feature로의 mapping function을 학습한다.
% 이 모델을 2초 관측 데이터로부터 6초 미래를 예측 하는 task인 nuScenes 데이터셋에서 평가하면 model은 과거 주행 feature를 미래 주행 feature로 제대로 mapping 하지 못한다.
% 이는 간단한 toy experiment에서 정량적으로 확인할 수 있다.
% 우리는 nuScenes의 train set을 1초 과거, 3초 미래, 10Hz로 임의로 interpolation을 수행했다.
% validation set은 원래대로 유지한 상태로 평가해보면 실험 결과와 같이 많은 예측 성능 하락이 있음을 확인하였고, time step configuration에 의한 domain discrepancy가 존재함을 확인하였다.
In doing so, we focus on two representative distinctions across datasets. 
First is the time step configuration difference, including observed/predicted time lengths and sampling frequencies as shown in Tab.~\ref{tab:intro_timestep_diff}.
% For many existing large-scale trajectory datasets, including Argoverse and WOMD, data is only accessible with fixed time length and frequency described in Tab.~\ref{tab:intro_timestep_diff}.
This results in the discrepancy of feature manifold of input/output trajectory in the feature space. 
For instance, a model trained on the WOMD dataset, which is to predict 8 seconds of future from 1 second of past with 10Hz, learns a mapping function between the past 1-second motion feature and the future 8-second motion feature. 
However, when evaluating this model on the nuScenes dataset which involves predicting 6 seconds into the future from 2 seconds of observed data in 2Hz, the model struggles to map past trajectory features to future ones accurately. 
% This discrepancy can be quantitatively observed through a straightforward toy experiment. 
% We conducted an arbitrary interpolation on the nuScenes train set, maintaining a 1-second past and 3-second future configuration at 10Hz. 
% Upon evaluation with the original nuScenes validation set (2/6s of history/future), we observed a significant decline in prediction performance, confirming the presence of domain discrepancies caused by varying time step configurations.

% 두번째, trajectory dataset은 차량에 부탁된 센서 데이터로부터 주변 차량을 detection/tracking 을 수행하여 얻은 데이터이다.
% 따라서 detection/tracking fault에 의해 data에 noise가 존재하며, 이 noise는 prediction 성능에 상당한 악영향을 끼친다는것이 알려져 있다~\cite{9811776, 9879091}.
% 한편, 취득한 데이터셋에 따라서 사용한 센서 종류, detector/tracker 종류가 다르기 때문에 이러한 tracklet error는 데이터셋마다 다른 양상으로 나타난다.
% 그림 ~~를 보면 nuScenes, Argoverse 데이터에 존재하는 noisy trajectory의 예시와 그로 인한 예측 실패 케이스를 나타낸다.
% nuScenes와 Argoverse 데이터에 포함된 tracklet noise는 time step configuration 에도 영향을 받는다.
% sampling rate에 따라 다른 noise 경향을 보이기 때문에 이러한 차이도 데이터셋의 discrepancy의 원인이 된다.

Secondly, trajectory datasets are obtained by detecting and tracking surrounding agents from the sensor data taken from the ego-agent. 
As a result, the tracked results (tracklets) are prone to both sensor noise and also inaccurate detection and tracking results~\cite{saleh2021probabilistic, park2020identifying}, and it adversely affects prediction performance~\cite{Weng2022_MTP}. 
Moreover, each dataset manifests unique tendencies of tracklet errors. 
It is because they use different types of sensors and detector/tracker configurations in the acquisition process. 
Their unique tendencies of tracklet errors are shown in Fig.~\ref{fig:intro_datadiff}. 
The tracklet noise is also influenced by the time step configuration, for different sampling rates exhibit unique noise patterns.
Namely, tracklet noise tends to be more severe with smaller $\Delta t$, as shown in Fig.~\ref{fig:intro_datadiff}, where Argoverse has more severe tracklet noise than nuScenes with the same past length.

% 우리는 이러한 차이를 다루기 위해 SDE의 continuous, stochastic representation을 사용한다.
% 기존에 discrete 하게 time series 데이터를 다루는 것 대신에, Neural Differential Equation 의 방식을 이용해 continuous 한 space에서 time series data를 다룬다.
% 또한 데이터셋에 다르게 존재하는 tracklet noise에 의한 uncertainty를 stochastic representation으로 handling할 수 있다는 것을 보여준다.
% 따라서 우리의 Contribution은 다음과 같다:

To address these disparities holistically, we adapt the continuous and stochastic representation of Neural Stochastic Differential Equation (NSDE). 
Rather than dealing with time series data discretely as conventional approaches, we leverage NSDE to handle time series data in a continuous space.
Additionally, we show the capability of stochastic representation in handling the tracklet errors.
Specifically, we propose a dataset-specific diffusion network of NSDE and its training method to enhance robustness against dataset-specific tracklet errors.
Our contributions are summarized as follows:

\begin{itemize}
    \item We utilize a continuous representation of NSDE for trajectory prediction to diminish internal discrepancies across datasets collected in arbitrary temporal configurations.
    \item We propose a framework of dataset-specific diffusion network and its training method to handle unique tracklet errors across datasets.
    \item The proposed methods are validated against state-of-the-art prediction methods including regression-based and goal-conditioned method, the two mainstreams of trajectory prediction methodology.
    \item We validate our methods across five datasets: nuScenes, Argoverse, WOMD, Lyft, and INTERACTION, and show consistent improvement in prediction accuracy with state-of-the-art prediction models.
\end{itemize}

\section{Related Works}

\subsection{Trajectory prediction}
% Trajectory prediction은 주변 차량의 미래 경로를 관측된 과거 경로와 HD map 정보와 같은 주변 환경 정보를 이용해 예측하는 task이다 (최근 방법 citations).
% 안전한 자율주행 시스템에 그것의 중요성이 매우 높아지면서, 최근에 여러 large-scale trajectory dataset 들이 구축되었다 (nuScnees, Argoverse, Waymo, Shift, Interaction, ...).
% 이 데이터셋은 주로 차량에 설치된 camera, lidar와 같은 센서로부터 주변 agent를 검출, tracking 하여 주행 경로를 취득한다.
% Interaction dataset같은 경우는 drone 을 이용해 얻어진 top-view 이미지로부터 경로를 얻는다.
% HD map 정보는 마찬가지로 센서 정보로부터 신호 정보를 받아 직접 검출할 수도 있고, 미리 구축된 HD map 으로부터 차선 정보를 얻기도 한다.
% 이러한 large-scale dataset들의 발표로, data driven 방식의 예측 모델의 성능이 갈수록 높아지고 있다.
% 차량간의 interaction을 잘 모델링하거나, 차량과 HD map 사이의 관계를 더 잘 capture 할 수 있는 방법이 다수 제안되었다.
% 또한 모델 구조적으로도 한번에 경로를 예측하는 regression-based model 뿐만 아니라, end-point를 따로 예측하고, 이를 conditioned 해서 주행 경로를 예측하는 goal-conditioned model 등 여러 구조의 모델이 제안되었다.
% 이러한 방법들은 모두 미리 구축된 large-scale 데이터셋에서 학습/평가가 이루어졌기 때문에 cross-domain 상황에서의 prediction은 아직 충분히 연구되지 않았다.

% ChatGPT 버전
Trajectory prediction involves predicting the future paths of road agents based on observed past trajectories and environmental information, such as HD maps~\cite{Wang2023WSiPWS, park2022leveraging}.% , aydemir2023adapt}. 
With its increasing interest, a number of large-scale trajectory datasets have been established~\cite{kothari2021human, malinin2021shifts}. 
These datasets acquire trajectories by detecting and tracking surrounding agents using sensor input installed on ego-agent. 
% In the case of the interaction dataset~\cite{interactiondataset}, trajectories are obtained from top-view images captured by drones. 
HD map information can be obtained from pre-built HD maps or derived from sensor data~\cite{hu2023_uniad}. 
The introduction of large-scale datasets has resulted in improved performance of data-driven trajectory prediction models. 
Various methods have been proposed to capture agent interactions or better relationship between HD maps~\cite{meng2022forecasting, salzmann2020trajectron++}. 
The methodology for motion forecasting based on these datasets' patterns could be broadly classified into two categories: regression-based and goal prediction-based models. 
Regression-based models predict the entire trajectory at once, while goal prediction-based models initially predict the endpoints, followed by conditional generation of motion path for each end points. 
However, despite the rapid advancements in the past few years, prediction in cross-domain scenarios remains relatively underexplored, as all these methods have been individually trained and evaluated on each pre-existing large-scale dataset.

\subsection{Cross-domain trajectory prediction}
% Trajectory dataset도 여러 데이터셋이 있는데 이 데이터셋 들간의 domain discrepancy가 있다는 것이 최근 논문에 의해 보여졌다~\cite{Gilles2022UncertaintyEF}.
% 그들은 nuScenes, Argoverse, Interaction, Shift 와 같은 여러 trajectory prediction 데이터셋들을 분석했는데, 데이터셋간의 transferability가 안좋다는 것을 확인하였다.
% 일반적인 domain adaptation 관점에서, 이러한 discrepancy를 해결하려는 방법이 제안되었다.
% ~~는 target dataset의 feature와 source dataset의 feature의 distribution distance를 줄이는 일반적인 domain adaptaion 방법을 제안하였다.
% ~~ 는 meta learning을 통한 offline/online adaptation 방법을 제안하였다.
% ~~ 는 hierarchical model 구조를 제안하여, domain specific 부분을 분리하여 domain adaptaion 방법을 제안하였다. 
% ~~ 는 federated learning 방법론을 이용하여 multi-source dataset에서 모델을 학습시켜서 모델 자체의 generality를 높이는 시도도 있었다.
% 한편, Trajectory dataset은 여러가지 요인에 의해 discrepancy가 존재한다.
% 우선, 데이터셋이 취득된 geological (external) factor에 따라 주행 환경이나 agent의 density 등이 달라져 주행 패턴이 달라질 수 있다.
% ~~는 그 중에 도로 구조의 curvature에 따른 discrepancy를 해결하고자 특별한 domain normalization 방법을 제안하였다.
% 그들이 제안한 Frenet 좌표계를 사용하면 일반적인 cartesian 좌표계에서는 컸던 도로 구조에 의한 discrepancy를 줄일 수 있었다.
% 그러나 앞서 서술한 방법 모두 time step configuration 차이에 따른 discrepancy를 고려하지는 않았다.
% domain adaptaion을 제안한 논문들은 모두 각자 cross-domain 실험을 세팅하였지만, 그들은 모두 time step을 도메인간의 겹치는 time step으로 통일하였다. (표현을 메끄럽게 수정 필요)
% 예를 들어 \cite{Gilles2022UncertaintyEF} 에서는 데이터셋이 공통적으로 포함하고 있는 과거 1초, 미래 3초의 time step configuration을 사용하였다.
% 또한 그들은 데이터셋이 수집될 때 detection/tracking error가 transferability에 영향을 미친다는 것을 지적하였지만, cross domain 환경에서 noise의 다른 tendency를 handling 하는 방법에대해서는 cover되지 않았다.

Recent research has highlighted the presence of domain discrepancies among various trajectory datasets~\cite{Gilles2022UncertaintyEF}. 
Analyzing datasets such as nuScenes, Argoverse, Interaction, and Shift, it has been confirmed that transferability between datasets is limited. 
From a general domain adaptation perspective, approaches have been proposed to address such discrepancies~\cite{9880042, Wang_2022_CVPR, wang2022transferable}. 
% One method (cite reference) suggests reducing the distribution distance between the feature representations of the target and source datasets as a common domain adaptation approach. Another approach (cite reference) proposes offline/online adaptation through meta-learning. Furthermore, a hierarchical model structure has been introduced (cite reference), separating domain-specific components to propose domain adaptation methods. Federated learning has also been explored as an attempt to enhance the generality of models by training them on multi-source datasets.
Besides, trajectory datasets exhibit discrepancies due to various factors. 
For instance, geographical (external) factors can lead to variations in driving environments and agent density, resulting in different driving patterns. 
To tackle such discrepancies related to road structure curvature, one method~\cite{Ye2023ImprovingTG} proposed a domain normalization technique using Frenet coordinates. 
% By using Frenet coordinates, the method could reduce discrepancies caused by differences in road structure observed in the conventional Cartesian coordinate system.
However, the methods mentioned above do not consider discrepancies arising from differenct data acquisition strategies including varing time step configuration and tracklet errors. 
While these domain adaptation papers all set up cross-domain experiments, they restricted the time steps to an overlap of all dataset time steps. 
For example, in \cite{Gilles2022UncertaintyEF}, they used a common time step configuration of 1 second past and 3 seconds future, which is shared by the datasets.
Additionally, although they acknowledged that detection/tracking errors during dataset collection could affect transferability~\cite{9811776, 9879091}, the different tendencies of error in cross-domain environments are yet to be addressed.

\subsection{Neural Differential Equation (NDE)}
% Neural ODE가 제안되었는데 continuous한 representation의 특징때문에 직관적으로 time series 데이터에 적용되었다.
% 주로 continuous 한 representation의 장점을 이용해 원해 함수를 예측하거나, video interpolation 과 같은 vision task에 이용되기도 하였다.
% encoder-decoder 구조로 모델링 하기도 하고 전체 time series의 latent를 continuous representation으로 표현하여 모델링하기도 하였다~\cite{, qian2022dcode}.
% time series latent를 곧바로 Neural DE로 표현하는 것은 high-dimensional feature를 표현하는데 도움이 되었지만 time series 데이터 외에 다른 modal 데이터를 같이 다루는 multi-modal data representation을 표현하는데는 어려움이 있다.
% 최근에는 NDE 를 이용해 Trajectory prediction을 푸려고 하는 시도들이 있었다.~\cite{10143287, 10.1007/978-3-031-20047-2_13}.
% Social ODE는 pedestrian trajectory prediction 문제에서 interaction 모델링이 가능하도록 Neural ODE를 도입하였다.
% MTP-GO 는 ~~~ 했다.
% 하지만 위 논문들은 neural DE의 continuous 한 특성을 trajectory prediction에 반영하지 않았고, stochasticity를 포함한 SDE 모델링과는 다르므로 본 논문에서 제안하는 contribution과는 큰 차이가 있다.
% ChatGPT 버전
The proposal of Neural Ordinary Differential Equations (NODE)~\cite{NEURIPS2018_69386f6b} has made significant strides in applying continuous representation to time series data, making it intuitive for various tasks like predicting continuous functions~\cite{anumasa2022latent, norcliffe2020neural} or other applications that need continuous time series representation~\cite{cao2023estimating, park2021vid}. 
Neural ODEs have been employed in encoder-decoder structures and have been used to represent the latent space of entire time series data in continuous form~\cite{
% rubanova2019latent, 
qian2022dcode}. 
% However, while directly applying Neural ODEs (NODE) to represent time series latents has been beneficial for expressing high-dimensional features, it poses challenges in handling multi-modal data representations which involve data from other modalities beyond time series information. 
Therefore, motion forecasting has been deemed as an epitome of a pattern recognition task solvable via NODEs for its temporally coordinated time series structure. 
The first work to utilize NODEs for motion forecasting was social ODE~\cite{10.1007/978-3-031-20047-2_13} which applied Neural ODEs to pedestrian trajectory prediction to enable interaction modeling. 
Moreover, the MTP-GO~\cite{10143287} constructed a graph-based NODE model for trajectory prediction.
% Moreover, the MTP-GO approach achieved significant progress in this domain. 
Nevertheless, these prior works have not fully leveraged the continuous characteristics of neural ODEs and lack the incorporation of stochastic nature of NSDE modeling, thereby our method substantially differs from previous frameworks based on NODEs.

\section{Method}

\begin{figure*}[t]
  \centering
    \includegraphics[width=1.0\linewidth]{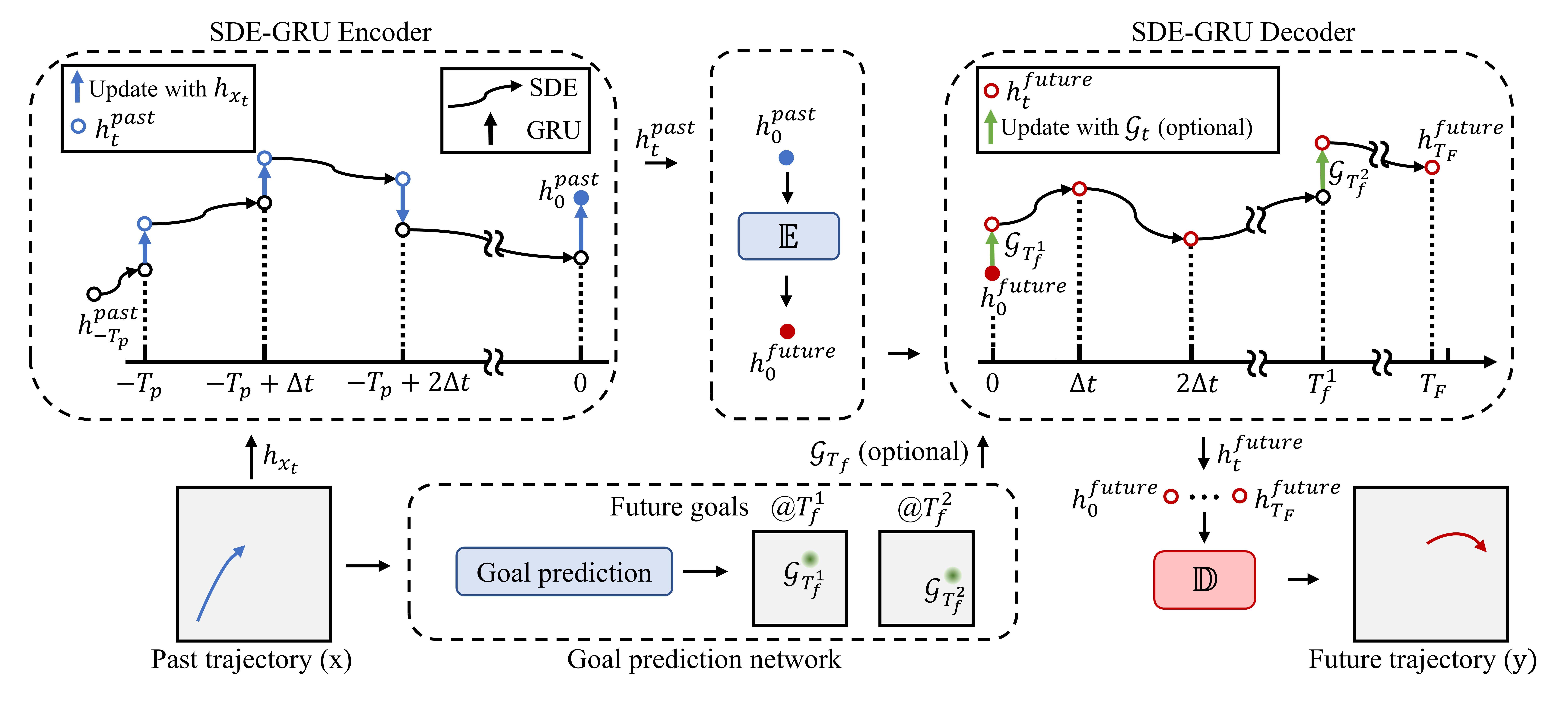}
  % \vspace{-5pt}
  \caption{Overall network architecture of the proposed SDE-GRU-based encoder-decoder model. Agent features are extracted from the input trajectory for all observed time steps $\textup{\textbf{h}}_{\textup{x}_t}$. SDE-GRU encoder then integrates the learnable parameter $h_{-T_p}^{past}$ from $-T_p$ to 0 along with feature updates via NSDE at unobserved time steps and GRU gating at observed past trajectory time steps. Followed by an additional encoding operation with encoder $\mathbb{E}$, the SDE decoder similarly integrates the encoded feature from time 0 to $T_F$. MLP decoder $\mathbb{D}$ then predicts the corresponding future motion.}
  % \vspace{-10pt}
  \label{method:model_architecture}
\end{figure*}

\subsection{Preliminaries}
\subsubsection{Problem definition}
% 일반적으로 Trajectory Prediction은 unique time step 의 과거 관측 trajectory {}와 semantic 정보 {}로 부터 미래 time step 동안의 trajectory를 예측하는 task이다.
Given $N$ agents, a position of a road agent $n \in \left \{ 1, ..., N\right \} $ at a specific time $t$ can be denoted as $\textup{\textbf{x}}_t^n$ for the past, and $\textup{\textbf{y}}_t^n$ for the future.
In general, trajectory prediction aims to predict future trajectory $\textup{Y} =$~\{ $\textup{\textbf{y}}_{\Delta t }^n, ... , \textup{\textbf{y}}_{T_f}^n$~\} from map information $M$ (optional) and observed history trajectory $\textup{X} =$~\{  $\textup{\textbf{x}}_{-T_p}^n, ... , \textup{\textbf{x}}_{-\Delta t }^n, \textup{\textbf{x}}_0^n$~\} with fixed time step $\Delta t$, history length $T_p$, and prediction horizon $T_f$.
In our problem definition, we assume that we have large-scale source dataset $\left \{ \textup{X}^{sr}, \textup{Y}^{sr} \right \}$ and small-scale target dataset $\left \{ \textup{X}^{tg}, \textup{Y}^{tg} \right \}$.
Because each dataset has own time step configuration ($T_p$, $T_f$, and $\Delta t$), we design a model that can handle arbitrary time-step sampled trajectory $\textup{Y} = \left \{ \textup{\textbf{y}}_t^n \right \}_{t \in (0, T_F]}$ and $\textup{X} = \left \{ \textup{\textbf{x}}_t^n \right \}_{t \in [-T_P, 0]}$ where $T_F$ and $T_P$ are maximum values of prediction horizon and history length across datasets.
From now, we omit the superscript $n$ for simplicity.

\subsubsection{Neural Stochastic Differential Equation}
% NODE는 hidden state h(t)의 initial value problem (IVP) 로 모델링 하고, 시간에 따른 feature의 transition을 neural net으로 모델링 한 방법이다.
% 그것의 continuous 한 특성때문에 irregularly sampled time series 데이터를 다루는데에는 NODE를 비롯한 NDE 기반의 방법들이 효과적이라고 알려져 있다~\cite{de2019gru, yildiz2019ode2vae, kidger2020neural}.
% 그 중 Neural Stochastic Differential Equation (NSDE)는 hidden state의 transition에 Brownian motion 의 term을 추가하여 stochasticity를 부여한 방법이다~\cite{tzen2019neural, liu2019neural}. 
% 이를 수식으로 나타내면 다음과 같다.
Neural Ordinary Differential Equation (NODE) is an approach that models the derivative of hidden state $\textup{\textbf{h}}_t$  employing neural networks to model the transition of features over time. 
Neural Stochastic Differential Equation (NSDE) introduces stochasticity by incorporating a term resembling Brownian motion into the transition of the hidden state~\cite{li2020scalable, tzen2019neural}. 
This can be represented as follows:
$$ d \textup{\textbf{h}}_t = f(\textbf{h}_t, t)dt + g(\textbf{h}_t, t)dW_t $$
% 여기서 W 는 standard Brownian motion 이며, f와 g 는 drift, diffusion function을 나타내며, neural network로 parametrized 된다.
% stochastic noise term이 data에 포함된 perturbation을 regularizer로써 완화시켜준다는 것이 알려져 있다.
Here, $W$ represents the standard Brownian motion, while $f$ and $g$ respectively denote the drift and diffusion functions and are parametrized by neural networks. 
The stochastic noise term acts as a regularizer, mitigating perturbations present in the data.
With the above derivatives, we can get a hidden state at a specific time t with initial value problem (IVP) solvers.

Thanks to its continuous nature across time, NDE is known as effective for handling irregularly sampled time series data~\cite{anumasa2022latent, kidger2020neural}. 
Therefore, we use NSDE to encode and decode temporal trajectory, which is originally performed with discrete networks like transformer~\cite{vaswani2017attention}, or LSTM~\cite{6795963} in previous methods.

\subsection{Proposed Framework}

\subsubsection{Modeling time-wise continuous latent}
Following conventions in both NDE and trajectory prediction, our model follows an encoder-decoder (sequence-to-sequence) structure as shown in Fig.~\ref{method:model_architecture}.
At first, we encode past trajectories of agents with SDE-GRU.
We adopt ODE-RNN structure to handle incoming irregularly sampled data, where ODE is replaced with SDE.
When the input positions is not observed at a time stamp, the latent is continuously translated via NSDE.
If the agent position at time t ($\textup{\textbf{x}}_t$) is observed, the latent vector ($\textup{\textbf{h}}_t$) is updated using encoded incoming data ($\textup{\textbf{h}}_{\textup{x}_t}$) via GRU following ~\cite{de2019gru, rubanova2019latent}.
How to obtain next step latent from current input and latent is as:
\begin{equation}
\begin{aligned}
& d \textup{\textbf{h}}_t = f(\textbf{h}_t, t)dt + g(\textbf{h}_t, t)dW_t  \\
& \textup{\textbf{h}}_{t+\Delta t}' = \textup{\textbf{h}}_t + f(\textup{\textbf{h}}_t, t) \Delta t + g(\textup{\textbf{h}}_t, t) \sqrt{\Delta t} W_t \\
& \textup{\textbf{h}}_{t + \Delta t} = \textup{GRU}(\textup{\textbf{h}}_{t+\Delta t}' , \textup{\textbf{h}}_{\textup{x}_t})
\end{aligned}
\label{eq:method_derivative}
\end{equation}
Here, $\textup{GRU}$ cell is represented as:
\begin{equation}
\begin{aligned}
& \textup{\textbf{r}}_t = \sigma ( W_r( \textup{\textbf{h}}_{t+\Delta t}' \oplus \textup{\textbf{h}}_{\textup{x}_t} )) + \textup{\textbf{b}}_r )  \\
& \textup{\textbf{z}}_t = \sigma ( W_z( \textup{\textbf{h}}_{t+\Delta t}' \oplus \textup{\textbf{h}}_{\textup{x}_t} )) + \textup{\textbf{b}}_z ) \\
& \textup{\textbf{g}}_t = \textup{tanh} ( W_g( (\textup{\textbf{r}}_t \odot  \textup{\textbf{h}}_{t+\Delta t}') \oplus \textup{\textbf{h}}_{\textup{x}_t} )) + \textup{\textbf{b}}_g ) \\
& \textup{\textbf{h}}_{t + \Delta t} = \textup{\textbf{z}}_t \odot \textup{\textbf{h}}_{t} + (1-\textup{\textbf{z}}_t) \odot \textup{\textbf{g}}_t
\end{aligned}
\label{eq:method_gru}
\end{equation}
where $\textup{\textbf{r}}_t$, $\textup{\textbf{z}}_t$, $\textup{\textbf{g}}_t$ correspond to reset gate, update gate, update vector and $\oplus$, $\odot$ correspond to concatenation, element-wise product.
Here, we omit the superscript $\small{\textit{past}}$ for Eqs.~\ref{eq:method_derivative},~\ref{eq:method_gru} for simplicity.
Then, integrating latent feature from $-T_p$ to 0, we get a latent feature of each agent at current time step ($t=0$):
\begin{equation}
\begin{aligned}
    & \textup{\textbf{h}}_0^{past} = \int_{-T_P}^{0} \textup{GRU}(\textup{SDEsolve}(\textup{\textbf{h}}_t^{past}, t), \textup{\textbf{h}}_{\textup{x}_t} ) dt \\
    & \textup{\textbf{h}}_0^{fut} = \mathbb{E}(\textup{\textbf{h}}_0^{past}, \mathcal{M})
\end{aligned}
\end{equation}
After encoding past trajectory as a single feature per agent, remaining part of encoder $\mathbb{E}$ is performed such as encoding with map information $\mathcal{M}$.

% The decoding process is analogous to that of the encoder.
% Here, we omit superscript $\small{\textit{fut}}$ for simplicity.
% NSDE is used to integrate the encoded feature ${\textbf{h}}_0$ from time 0 to $T_F$. 
% The main difference compared to the encoder is the absence of GRU-based hidden state update at each timestep because there is no incoming data during the decoding stage. 
% Instead, hidden states at all timesteps $\left \{ \textup{\textbf{h}}_t \right \}_{t \in (0, T_F]}$ are directly fed into the decoder ($\hat{\mathbf{y}}_t = \mathbb{D}(\textup{\textbf{h}}_t)$).
% Compared to the encoder where only the final hidden state $h_0$ has been extracted, an MLP decoder extracts future position $\hat{\textup{Y}} = \left \{ \hat{\mathbf{y}}_t \right \}_{t \in (0, T_F]}$ from all decoded hidden states by their corresponding timestep. 

In case of decoder, it has different network design depending on whether the base model is regression-based model or goal-conditioned model.
Unlike the past feature which needs to be updated as data coming as t passes, there is no incoming data for future decoding.
Therefore, in regression-based model, hidden state at future time step can be obtained by vanilla SDE solver without GRU update.
However, in case of goal-conditioned method, we propose a multi-scale-goal updating method as depicted in the right part of Fig~\ref{method:model_architecture}. 
The goal-conditioned decoder predicts trajectory both from $\textup{\textbf{h}}_0$ and goal feature. 
Goal is predicted at the last time step for each dataset configuration: $\pmb{\mathcal{G}}_{T_f^1}$, $\pmb{\mathcal{G}}_{T_f^2}$.
Here, $T_f^1$ is a time step which is smaller one between \{ $T_f^{sr} , T_f^{tg}$ \}, and $ T_f^2 $ is the larger one which is identical with $T_F$.
To adopt this multi-scale goal-conditioned, we additionally utilize SDE-RNN which can be represented as:
\begin{equation}
    \textup{\textbf{h}}_{t \in \left ( 0, T_F \right ] } = \left\{\begin{matrix}
\begin{aligned} 
& \textup{SDEsolve} ( \textup{GRU} ( \textup{\textbf{h}}_0, \pmb{\mathcal{G}}_{T_f^1} ) ) &  0 < t \leq T_f^1
\\
& \textup{SDEsolve} ( \textup{GRU} ( \textup{\textbf{h}}_{T_f^1}, \pmb{\mathcal{G}}_{T_f^2} ) ) &  T_f^1 < t \leq T_F
\end{aligned} 
\end{matrix}\right.
\end{equation}
Finally, a mlp decoder $\mathbb{D}$ is utilized to decode future position $\hat{\textup{Y}} = \left \{ \textup{\textbf{x}}_t \right \}_{t \in (0, T_F]}$ from hidden state $\left \{ \textup{\textbf{h}}_t \right \}_{t \in (0, T_F]}$.

\subsubsection{Handling tracklet uncertainty}
% is inevitably incorporated within the trajectory data since it is obtained via detection and tracking from sensor data. 
% Their tendencies are unique  
% One significant cause to such dataset-specific noise is the time step configuration during data acquisition. 
% Namely, tracklet noise tends to be more severe with smaller $\Delta t$ during data acquisition. 
% % With compared to smaller $\Delta t$, tracklet noise tend to be severe with smaller $\Delta t$.
% Therefore, even with the use of a neural representation of differential equation which is known to be effective in handling data perturbation, such competence on one dataset is unlikely to be reproduced on other datasets.

\begin{figure}[t]
  \centering
    \includegraphics[width=.99\columnwidth]{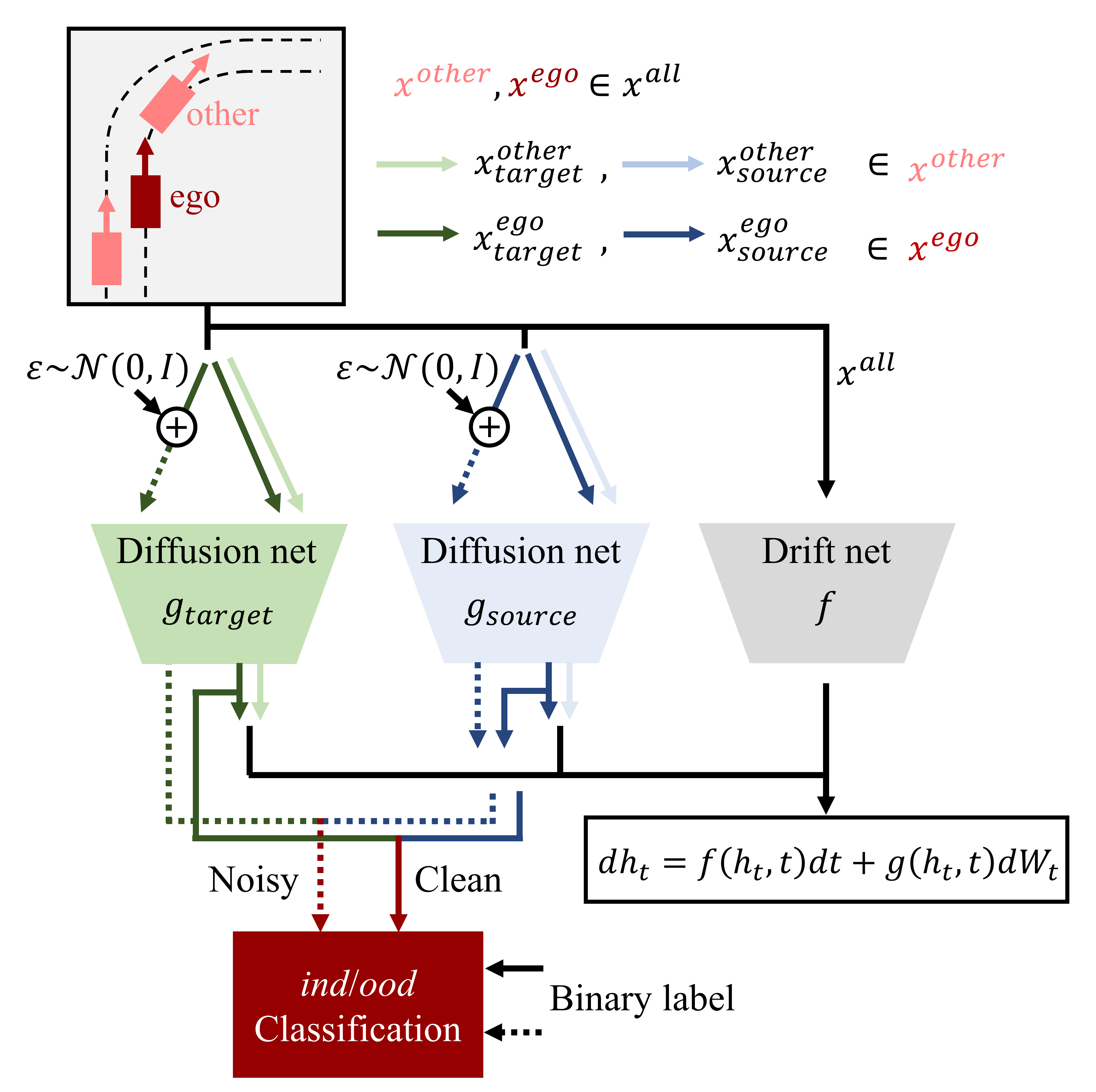}
  % \vspace{-5pt}
  \caption{Visual representation of the proposed uncertainty training framework. We use a shared drift net and separated diffusion nets for source and target datasets. Unlike non-ego agents, ego-agents' trajectories with noise perturbation are additionally fed to the diffusion nets. The outputs from both drift networks and diffusion networks are utilized to solve the stochastic differential equation. In addition, diffusion networks are separately trained to perform in-distribution (\textit{ind}) and out-of-distribution (\textit{ood}) classification of ego-agent trajectories with and without perturbation. As a result, brownian motion $dW_t$ takes larger weight on the subsequent SDE integration for uncertain samples.}
  % \vspace{-5pt}
  \label{method:detailed_sde}
\end{figure}

As elaborated in previous sections, the tendencies of tracklet error are unique across datasets.
Although the NSDE is known to be robust to data perturbation, it is troublesome to account for each and every uncertainty tendency at once.
With this motivation, we adopt the concept of SDE-Net~\cite{kong2020sde}.
% SDE-Net은 in-distribution data (ind) 와 out-of-distribution data (ood) 를 이용하여 diffusion network를 학습시킨다.
% 그들은 SDE derivative 계산식에서 diffusion net을 학습시킬 때 ind에 대해서 0을, ood 에 대해서 1로 가도록 학습한다.
% 이를 이용해 학습이 되지 않은 ood data에 대해 model이 uncertainty를 측정할 수 있는 방법을 제안하였다.
% 우리가 제안하는 방법은 두가지 측면에서 다르다. 
% 첫번째로 우리는 out-of-distribution data를 다루는 것이 아니라, trajectory data에 포함된 tracklet noise을 검출하여 regularizer로 사용할 수 있도록 한다.
% 이를 위해 우리는 ind를 clean trajectory data, ood를 noise가 포함된 trajectory data로 설정한다.
%    $$ Equation $$
% 이 때, 데이터셋에 존재하는 agent 중에 tracklet noise가 포함되지 않은 clean trajectory data는 ego-agent 의 경로를 사용한다.
% 이는 ego-agent의 경로는 sensor input에서 얻어진 것이 아니라 GPS와 localization 정보로 얻어진 것이기 때문에, tracklet noise가 가장 흔하게 발생하는 occlusion이나 id switch로부터 자유롭기 때문이다
% ood 데이터는 ego-agent의 경로에 Gaussian noise를 더해서 만든다.
% 이렇게 함으로써, sde encoder는 trajectory data의 tracklet noise가 커서 uncertainty가 크게 측정된 샘플에 대해서는 diffusion noise, 즉, regularize 효과를 줄 수 있다.
% 두 번째는 데이터셋마다 다른 tracklet noise를 handling 하기 위해 데이터셋에 따라 다른 diffusion network를 사용한다.
% NSDE formulation에서 drift net은 system control을 통해 좋은 prediction accuracy를 달성하는데 있고, diffusion net은 aleatoric uncertainty를 모델링 하는데 있다.
% 따라서 cross dataset의 learning의 prediction accuracy 효과를 enjoy 하기 위해 우리는 drift net은 데이터셋간에 공유하고, diffusion network는 따로 쓰는 모델을 설계하였다. 
% 최종적으로, 우리가 제안하는 모델 식과 학습 objective는 다음과 같다:
The SDE-Net utilizes both in-distribution data (\textit{ind}) and out-of-distribution data (\textit{ood}) to train the diffusion network. They accomplish this by training the diffusion net to assign 0 to \textit{ind} and 1 to \textit{ood} in the SDE derivative computation. 
With this approach, the model accomplishes two things: 1. Uncertainty measurement of previously unseen \textit{ood} data, 2. Larger weighting of brownian motion term $dW_t$ for uncertain samples when solving SDE. 
Its objective is the following:
\begin{equation}
     \min _{\boldsymbol{\theta}_g} \mathrm{E}_{\textup{\textbf{h}}_t \sim P_{\textit{ind}}} g\left(\textup{\textbf{h}}_t , t \right ) + \max _{\boldsymbol{\theta}_g} \mathrm{E}_{\tilde{\textup{\textbf{h}}}_t \sim P_{\textit{ood}}} g\left(\tilde{\textup{\textbf{h}}}_t , t \right )
\end{equation}

Compared to the SDE-Net, our method differs in two aspects. 
Firstly, we focus on detecting and utilizing trajectory samples with tracklet error within a dataset, rather than addressing cross-dataset distribution. 
To this end, we define \textit{ind} as a clean trajectory and \textit{ood} as trajectory containing tracklet error. 
During training, the ego-agent's trajectory is used for clean data since it is obtained from GPS and localization information, thus free from tracklet errors originating from occlusion or ID switch. 
For \textit{ood} data, Gaussian noise is added to the ego-agent's trajectory. 
It is based on the fact that most multi-object trackers are based on recursive Bayesian filters, so they produce Gaussian state uncertainty ~\cite{9811776}.
This way, the encoder NSDE is trained to assign larger weight on Brownian motion term $dW_t$ for noisy trajectory data, and smaller weight for clean data.
Since the Browninan motion term of NSDE is known to act as a regularizer~\cite{liu2019neural}, our method fosters robustness by intensifying regularization effect on noisy trajectory samples through diffusion network-driven Brownian motion weighting.
% induces a more robust model by assigning more regulaizing effect on noisy trajectory data sample via diffusion network-based Brownian motion term weighting.

Secondly, we address the tracklet error variation across datasets by using separate diffusion networks per dataset. 
In the NSDE formulation, the drift net aims to achieve high prediction accuracy via system control, while the diffusion net captures aleatoric uncertainty. 
To leverage multi-source learning for prediction accuracy, we share the drift net across datasets and allocate distinct diffusion networks for each dataset.
Ultimately, the derivative of NSDE encoder and its training objective is formulated as follows: 
\begin{equation}
    d \textup{\textbf{h}}_t = f(\textbf{h}_t, t)dt + g(\textbf{h}_t, t)dW_t , \quad  g=\left\{\begin{aligned}
    g_{sr} & \ \text{if} \ \textup{X}^{sr} &  \\
    g_{tg} & \ \text{if} \ \textup{X}^{tg} &  \\
    \end{aligned}\right.
\end{equation}
\begin{equation}
\begin{aligned}
    \min _{\boldsymbol{\theta}_{g_{sr}}} \mathrm{E}_{ \textup{\textbf{h}}_t \sim P_{ego}^{sr}} & g\left(\textup{\textbf{h}}_t , t \right ) + \max _{\boldsymbol{\theta}_{g_{sr}}} \mathrm{E}_{\tilde{\textup{\textbf{h}}}_t \sim P_{ego}^{sr}} g\left( \tilde{\textup{\textbf{h}}}_t , t \right ) \\
    + & \min _{\boldsymbol{\theta}_{g_{tg}}} \mathrm{E}_{\textup{\textbf{h}}_t \sim P_{ego}^{tg}} g\left(\textup{\textbf{h}}_t , t \right ) + \max _{\boldsymbol{\theta}_{g_{tg}}} \mathrm{E}_{\tilde{\textup{\textbf{h}}}_t \sim P_{ego}^{tg}} g\left( \tilde{\textup{\textbf{h}}}_t , t \right )
\end{aligned} 
\label{eq:final_g_obj}
\end{equation}
Here, $g_{sr}$ and $g_{tg}$ are diffusion networks for source and target dataset, respectively.
$\textup{\textbf{h}}_t$ and $\tilde{\textup{\textbf{h}}}_t$ corresponds to hidden states encoded from clean ego-agent trajectories and noise-injected ego-agent trajectories. Figure.~\ref{method:detailed_sde} illustrates the proposed uncertainty training framework.

\subsubsection{Training details}
In addition to the objective of Eq.~\ref{eq:final_g_obj}, prediction loss between $\hat{\textup{Y}}$ and GT $\textup{Y}$ is also used in order to train the overall encoder-decoder structured model. 
In detail, the prediction loss trains the decoder SDE to learn a continuous transition function between the encoded feature $\textbf{h}_0$ and future timestep hidden states $\left \{ \textup{\textbf{h}}_t \right \}_{t \in (0, T_F]}$ and encoder SDE to produce a continuous representation of the observed trajectory $X$, thus enabling the model to handle arbitrary time lengths and steps.
Additionally, if the baseline model is a goal-conditioned model, goal prediction loss between predicted goal position $\hat{\mathcal{G}}$ and gt goal $\mathcal{G}$ is used.
For the implementation of SDE methods, \textit{torchsde}~\cite{torchsde_git} is used.
For the initial latent value ($\textup{\textbf{h}}_{-T_p}^{past}$), we assign a learnable parameter.
Please refer to the supplementary materials for additional details on the training procedure and model implementation. 

\section{Experiment}
\subsection{Experiment setup}
Our goal is enhancing performance in target dataset via additional training on a large-scale source dataset.
To assess whether the prediction model alleviates limited transferability across datasets, we jointly train with training set of both dataset and evaluate on validataion set of target dataset.
In detail, the regression-based prediction model is trained only with nuScenes \textit{train} set (\textbf{N}), jointly with nuScenes + Argoverse \textit{train} sets (\textbf{N+A}) or nuScenes + WOMD \textit{train} sets (\textbf{N+W}), then validated on nuScenes \textit{val} set, with a widely used metric, mADE$_{10}$.
To evaluate the model's generalizability on different target datasets, we additionally report training on WOMD.
As for the goal-conditioned model, its improvement on nuScenes and Lyft validation sets are evaluated.
Specifically, for nuScenes validation, we compare training only on nuScenes \textit{train} set (\textbf{N}) and jointly with nuScenes + INTERACTION \textit{train} sets (\textbf{N+I}).
We additionally report its improvement on Lyft validation set by comparing only training on Lyft \textit{train} set (\textbf{L}) and jointly with Lyft + INTERACTION \textit{train} sets (\textbf{L+I}).

To enable discrete baseline models to be applied to two different time-step configured data, we have re-arranged both trajectory datasets.
We create empty time series data that can contain both dataset time steps, then scatter each data to each time step.
For example, with nuScenes (2/6s, 2Hz) as target and Argoverse (2/3s, 10Hz) as source, we create 81 bins (2/6s, 10Hz) of empty data.
Then, nuScenes data is scattered with 5 time step intervals for overall lengths, while only the first 50 time steps are filled for Argoverse data.

\subsection{Datasets and baseline model}
We use nuScenes (30k) and Lyft (160k) as small-scale target dataset for their relatively smaller sizes.
We utilize INTERACTION, Argoverse (200k) and WOMD (500k) datasets as largse-scale datasets for additional training.
To utilize common information among datasets, we use past/future trajectories and lane centerline information only.
Additionally, while these datasets have both vehicle and pedestrian trajectory data, we only train and evaluate vehicle trajectories for simplicity.
To show the effectiveness of our framework, we select HiVT~\cite{Zhou_2022_CVPR} and MUSE-VAE~\cite{Lee_2022_CVPR} as the latest regression and goal prediction-based trajectory prediction method and show that even the state-of-the-art method has room for improvement with the fusion of our proposed SDE framework.
% To show the effectiveness of our framework, we select the latest regression-based trajectory prediction method HiVT~\cite{Zhou_2022_CVPR} as baseline regression-based model and MUSE-VAE~\cite{Lee_2022_CVPR} as baseline goal prediction-based model and show that even the state-of-the-art method has room for improvement with the fusion of our proposed SDE framework.

% HiVT currently achieves state-of-the-art performance on the argoverse dataset, and we have selected HiVT to show that even the state-of-the-art method has room for improvement with the fusion of our proposed SDE framework.

% \subsection{Baseline models}
% To show the effectiveness of our framework, we selected the latest regression-based trajectory prediction method HiVT~\cite{Zhou_2022_CVPR} as baseline model. HiVT currently achieves state-of-the-art performance on the argoverse dataset, and we have selected HiVT to show that even the state-of-the-art method has room for improvement with the fusion of our proposed SDE framework.

\section{Results}
\subsection{Effectiveness in multi-source training}
\label{result:quantitative}
\begin{table}[t]
\centering
\small\begin{tabular}{l|c|cc|cc} \toprule
                       % & \multicolumn{5}{c}{$\textup{mADE}_{10}$ on nuScenes \textit{val} set}                                                                    \\ \cmidrule{2-6}
Train Set                            & \textbf{N}       & \textbf{N}+\textbf{A}            & {gain}               & \textbf{N}+\textbf{W}            & gain             \\ \hline %\cmidrule(lr){2-2} \cmidrule(lr){3-4} \cmidrule(lr){5-6}
\small{\makecell[c]{vanilla \\ HiVT}}           & 1.045     & 0.966          & 7.56\%               & 0.950          & 9.09\%           \\ \midrule
\small{\makecell[c]{HiVT + \\ ODE-RNN}}         & 1.058     & 0.935          & 11.62\%              & 0.913          & 13.71\%          \\ \midrule
\small{\makecell[c]{HiVT + \\ latentSDE}}       & 1.044     & 0.943          & 9.67\%               & 0.912          & 12.64\%          \\ \midrule
\textbf{\small{\makecell[c]{HiVT + \\ ours}}}            & 1.044     & \textbf{0.913} & \textbf{12.55}\%     & \textbf{0.893} & \textbf{14.46}\% \\ \bottomrule
\end{tabular}
\caption{Effectiveness on regression-based method. All digits represent mADE$_{10}$ on nuScenes \textit{val} set trained on each dataset and the corresponding gain. \textbf{N}, \textbf{A}, and \textbf{W} respectively denote nuScenes, Argoverse, and WOMD. Lower is better.}
\label{tab:result_multi_source_hivt}
\end{table}

% \begin{table*}[t]
% \centering
% \small\begin{tabular}{l|c|cc|cc} \toprule
%                        % & \multicolumn{5}{c}{$\textup{mADE}_{10}$ on nuScenes \textit{val} set}                                                                    \\ \cmidrule{2-6}
% Train Set                            & nuS \textit{train}       & \makecell{nuS \textit{train} \\ +Argo \textit{train}}            & {gain}               & \makecell{nuS \textit{train} \\+WOMD \textit{train}}            & gain             \\ \hline %\cmidrule(lr){2-2} \cmidrule(lr){3-4} \cmidrule(lr){5-6}
% \small{\makecell[c]{vanilla \\ HiVT}}           & 1.045     & 0.966          & 7.56\%               & 0.950          & 9.09\%           \\ \midrule
% \small{\makecell[c]{HiVT + \\ ODE-RNN}}         & 1.058     & 0.935          & 11.62\%              & 0.913          & 13.71\%          \\ \midrule
% \small{\makecell[c]{HiVT + \\ latentSDE}}       & 1.044     & 0.943          & 9.67\%               & 0.912          & 12.64\%          \\ \midrule
% \textbf{\small{\makecell[c]{HiVT + \\ ours}}}            & 1.044     & \textbf{0.913} & \textbf{12.55}\%     & \textbf{0.893} & \textbf{14.46}\% \\ \bottomrule
% \end{tabular}
% \caption{Effectiveness on regression-based method. All digits represent mADE$_{10}$ on nuScenes \textit{val} set trained on each dataset and the corresponding gain. \textbf{N}, \textbf{A}, and \textbf{W} respectively denote nuScenes, Argoverse, and WOMD. Lower is better.}
% \label{tab:result_multi_source_hivt}
% \end{table*}

%
Table.~\ref{tab:result_multi_source_hivt} shows the improvements due to multi-source training on the regression-based model.
We compare our method with original discrete models, as well as ODE-RNN~\cite{rubanova2019latent} and LatentSDE~\cite{li2020scalable} adaptations.
% In the first column, the baseline model is trained with \textbf{N}, and with \textbf{N+A} in the second column (Argoverse) and third column (WOMD).
Compared to training with \textbf{N}, the baseline model's mADE has improved 7.56\% for \textbf{N+A}, and 9.09\% for \textbf{N+W}.
This improvement signifies an underfitted result when the model is only trained with nuScenes, a relatively smaller dataset that is comprised of only 30k training data. 
However, further performance gain has been limited since the discrete temporal encoding of vanilla HiVT is incapable of efficiently handling the cross-dataset discrepancy. 
By adopting ODE-RNN as the temporal encoder/decoder, the use of additional training data has brought about much more performance improvement (11.62\% for \textbf{N+A}, 13.71\% for \textbf{N+W}) thanks to its continuous modeling of latent transition across time.
Although adopting latentSDE is known to be robust against data perturbation, its performance gain slightly decreases compared to ODE-RNN (9.67\% for \textbf{N+A} and 12.64\% for \textbf{N+W}).
It is because the single diffusion network of latentSDE failed to address different type of tracklet error across datasets.
Finally, with our proposed method, mADE$_{10}$ improves to 0.913 for \textbf{N+A}, and 0.893 for \textbf{N+W}.
These improvements correspond to 12.55\%/14.46\% compared to nuScenes \textit{only} training (\textbf{N}), and 5.49\%/6\% compared to vanilla HiVT (0.966 $\to$ 0.913 and 0.950 $\to$ 0.893) which empirically show the effectiveness of the proposed methods.

Table.~\ref{tab:result_goal_conditioned} shows the effectiveness of our method on the goal-conditioned model.
We conduct experiments on two different target validation datasets: \textbf{N} and \textbf{L}.
% , Tab. reports experiments on two different validation set: nuScenes and Lyft.
For each case, we compare the performance gain when using \textbf{I} set as additional training data. 
The use of our method resulted in significant improvement in performance gain compared to the vanilla MUSE-VAE model.  
Specifically, its performance gains on both validation sets are threefold compared to the vanilla MUSE-VAE method, demonstrating the importance of improved transferability of our method across different types of backbone prediction models.
In addition, similar to the goal-conditioned model's generalized improvement on two different target sets, the regression-based method also shows improvement in performance on a different target validation set. 
Table~\ref{tab:result_hivt_WOMD_target} reports the regresssion-based method's performance on \textbf{W} set as target set.
Use of only 5\% of \textbf{W} set is compared to additional use of \textbf{A} set to assume a situation where only a small amount of training data within target set distribution is available. 
The use of our method over the baseline HiVT again shows significant improvement in performance gain.

% \begin{table}[]
% \centering
% \resizebox{\columnwidth}{!}{%
% \begin{tabular}{cccc}
% \hline
% Training set                        & N                          & N+I   & gain   \\ \hline \hline
% \multicolumn{1}{c|}{museVAE}        & \multicolumn{1}{c|}{2.304} & 2.178 & 5.47\%  \\ \hline
% \multicolumn{1}{c|}{museVAE + Ours} & \multicolumn{1}{c|}{2.333} & 1.953 & 16.29\% \\ \hline
% \end{tabular}%
% }
% \end{table}

% \begin{table}[t]
% \centering
% \resizebox{\columnwidth}{!}{%
% \begin{tabular}{c|ccc|ccc}
% \toprule
% Val. set                                           & \multicolumn{3}{c|}{NuS}                    & \multicolumn{3}{c}{Lyft}                    \\ \hline 
% Train set                                             & \multicolumn{1}{c|}{N}     & N+I   & gain   & \multicolumn{1}{c|}{L}     & L+I   & gain   \\ \hline \hline
% museVAE                                                  & \multicolumn{1}{c|}{2.304} & 2.178 & 5.47\%  & \multicolumn{1}{c|}{1.179} & 1.073 & 8.90\%  \\ \hline
% \begin{tabular}[c]{@{}c@{}}\textbf{museVAE +}\\ \textbf{Ours}\end{tabular} & \multicolumn{1}{c|}{2.333} & \textbf{1.953} & \textbf{16.29\%} & \multicolumn{1}{c|}{1.191} & \textbf{0.827} & \textbf{30.56}\% \\ \bottomrule
% \end{tabular}%
% }
% \caption{Effectiveness on goal-conditioned method. All digits represent mADE$_{10}$ on nuScenes and Lyft \textit{val} set trained on each dataset and the corresponding gain. The lower is better.}
% \label{tab:result_goal_conditioned}
% \end{table}

\begin{table}[t]
\centering
\scriptsize
\resizebox{\columnwidth}{!}{%
\begin{tabular}{c|c|c|c}
\toprule
\multicolumn{1}{c|}{Valid set}    & \multicolumn{1}{c|}{Train set}      & \multicolumn{1}{c|}{museVAE} & \multicolumn{1}{c}{\makecell{museVAE + Ours}} \\ \toprule
\multirow{3}{*}{\textbf{N}}                & \textbf{N}                              & 2.304                        & 2.333                               \\ %\cline{2-4} 
                                        & \makecell{\textbf{N+I}}                & 2.178                        & \textbf{1.953}                               \\ \cline{2-4} 
                                        & gain                                   & 5.47\%                       & \textbf{16.29\%}                             \\ \hline
\multirow{3}{*}{\makecell{\textbf{L}}} & \makecell{\textbf{L}}               & 1.179                        & 1.191                               \\ %\cline{2-4} 
                                        & \makecell{\textbf{L}+\textbf{I}} & 1.073                        & \textbf{0.827}                               \\ \cline{2-4} 
                                        & gain                                   & 8.90\%                       & \textbf{30.56\%}                             \\ \bottomrule
\end{tabular}
}
% \vspace{-5pt}
\caption{Effectiveness on goal-conditioned method on two different target datasets. All digits represent mADE$_{10}$ on nuScenes and Lyft \textit{val} set trained on each dataset and the corresponding gain. \textbf{N}, \textbf{L}, and \textbf{I} respectively denote nuScenes, Lyft, and INTERACTION. The lower is better.}
% \vspace{-5pt}
\label{tab:result_goal_conditioned}
\end{table}

\begin{table}[t]
\centering
\scriptsize
\resizebox{\columnwidth}{!}{%
\begin{tabular}{c|c|c|c}
\toprule
\multicolumn{1}{c|}{Valid set} & \multicolumn{1}{c|}{Train set}       & \multicolumn{1}{c|}{HiVT} & \multicolumn{1}{c}{\makecell{HiVT + Ours}} \\ \toprule
\multirow{3}{*}{\makecell{\textbf{W}}}       & \makecell{\textbf{W} (5\%)}                     &  0.9454                         & 0.9445                                 \\ %\cline{2-4}
                                & \makecell{\textbf{W} (5\%) + \textbf{A}} & 0.9286                          & \textbf{0.8496}                                 \\ \cline{2-4} 
                                & gain                                 &  1.78\%                         & \textbf{10.05\%}                                 \\ \bottomrule
\end{tabular}
}
\caption{Additional experiment of regression-based method on WOMD validation set as target dataset, again showing effectiveness on different target dataset. All digits represent mADE$_{10}$ on nuScenes \textit{val} set trained on each dataset and the corresponding gain. \textbf{W} and \textbf{A} respectively denote Waymo and Argoverse. The lower is better.}
\label{tab:result_hivt_WOMD_target}
\end{table}

% \begin{table}[t]
% \small\begin{tabular}{lccccc} \toprule
%                        & \multicolumn{5}{c}{$\textup{mADE}_{10}$ on nuScenes \textit{val} set}                                                                    \\ \cmidrule{2-6}
% Training Set           & {N}     & N+A            & {gain}             & N+W            & gain             \\ \cmidrule(lr){2-2} \cmidrule(lr){3-4} \cmidrule(lr){5-6}
% \makecell[l]{vanilla \\ MUSEVAE}           & - & -         & -           & -          & -           \\ 
% \makecell[l]{MUSEVAE + \\ ODE-RNN}         & - & -          & -          & -          & -          \\ 
% \makecell[l]{MUSEVAE + \\ latentSDE}       & - & -          & -           & -          & -          \\ 
% \makecell[l]{MUSEVAE + \\ ours} & - & - & - & - & - \\ \bottomrule
% \end{tabular}
% \caption{Effectiveness on goal-conditioned method. All digits represent mADE$_{10}$ on nuScenes \textit{val} set trained on each dataset and corresponding gain. The lower is better.}
% \label{tab:my-table_}
% \end{table}

\subsection{Effect of target dataset size}
\label{result:size}
\begin{figure}[t]
  \centering
  % \hspace*{-0.038\linewidth}
  \includegraphics[width=\columnwidth]{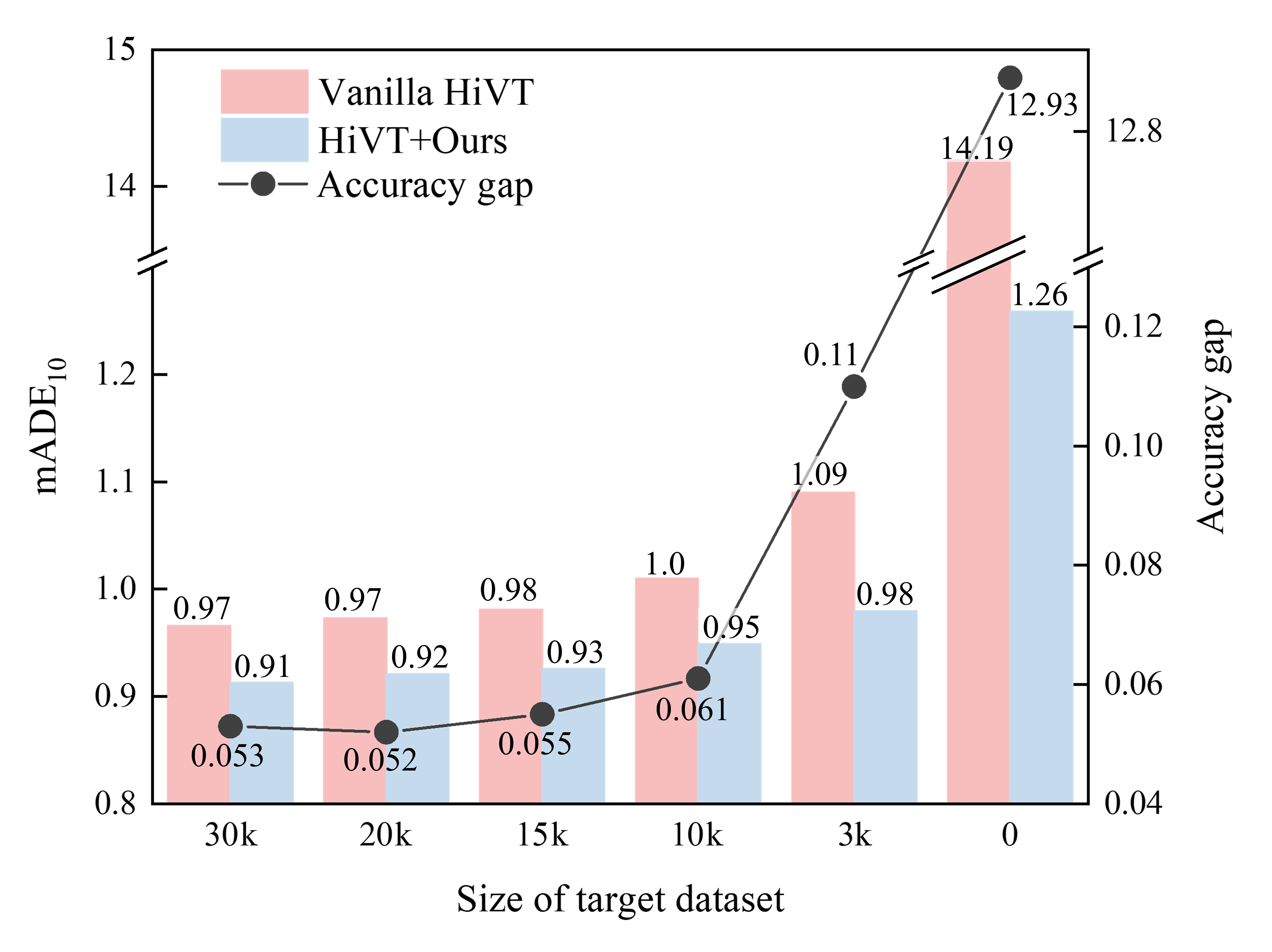}
  % \vspace{-20pt}
  \caption{Prediction accuracy according to the size of the target dataset. The target dataset is nuScenes, and the source is Argoverse. X-axis denotes the amount of nuScenes training set used during training. The Y-axis bars indicate mADE$_{10}$ of both models, and the grey line indicates the accuracy gap between baseline HiVT and the proposed method.}
  \label{fig:result_size}
\end{figure}

Previous experiments have been conducted with the size of target dataset as 30k, the size of nuScenes \textit{train} set.
However, 30k is still a considerably large-scale dataset, and collecting a labeled dataset of an equivalent size could still be considered a cumbersome work.
% Therefore, we have also conducted the same experiments with smaller sizes of target dataset to show the effectiveness of our model when only a smaller target dataset is available.
Therefore, we have also conducted the same experiments with smaller sizes of target dataset to show the effectiveness of our model even when the available target dataset is smaller.
By randomly dropping a ratio of nuScenes \textit{train} set, we construct the target datasets size of 20k, 15k, 10k, 3k, and 0 (no target dataset is used).
Argoverse dataset is used as the source dataset.
In the case of 0 target dataset setting, we share diffusion net and maintain the uncertainty training objective in Eq.~\ref{eq:final_g_obj}.
In Fig.~\ref{fig:result_size}, mADE$_{10}$ of the baseline and our method are plotted with bar graph, and their difference is plotted as line graph.
The effectiveness of the proposed method gradually increases in the range of 30k to 15k, and exponentially increases with the smaller target dataset sizes.
% This result shows that our proposed method is more effective as the size of target dataset is smaller thanks to the improved transferability across datasets.
Such larger improvements on smaller target datasets show that our proposed method effectively promotes transferability across datasets.
Indeed, at an extreme with no target dataset for training, mADE of the baseline diverges over 14 while ours remain reasonable at 1.26. 
% These results show that our proposed NDE temporal encoder effectively handles the cross-dataset discrepancy due to time step configuration difference.
These results show that our proposed NSDE temporal networks's advantage of effective cross-dataset discrepancy handling is even more valued for smaller target datasets.

\subsection{Effectiveness of continuous representation}

We compare our NSDE with the baseline HiVT model equipping with other methods for handling dataset-wise unique time step configuration as reported in Tab.~\ref{tab:result_time_aug}.
% We conducted an experiment to show the effectiveness of continuous representation of NSDE across time in Tab.~\ref{tab:result_time_aug}.
% To show its validity, we compared our method with other methods that could diminish discrepancy due to time step configuration difference.
First method is random dropping (RD) where some portion of time steps is randomly dropped during training.
We expect RD to be equivalent to stochastic noise injection, thus improving generalizability.
However, RD shows minimal improvement of only 0.006 since dropping time steps does not provide any extra time step data to a discrete temporal network.
% Besides, use of RD for time series data is limited since manipulation of data's time step could be simply done via interpolation or extrapolation.
% Besides, we can interpolate or extrapolate time series data to manipulate it to another time step configuration.
In that sense, we experiment with manipulating source data (1/8s, 10Hz) to target data's time step configuration (2/6s, 2Hz) (\textbf{S $\to$ T}) or vise-versa (\textbf{T $\to$ S}) through interpolation and extrapolation.
% We can manipulate source data to have same configuration with target data (\textbf{S $\to$ T}) or vise-versa (\textbf{T $\to$ S}).
Converting target dataset to source dataset configuration severely downgrads the prediction performance due to the source data's inaccurate information obtained from extreme extrapolation to unseen future time steps.
While converting source dataset to target dataset slightly increases accuracy, its improvement remains minimal due to inaccurate extrapolation of past trajectory.
% In addition, a knowledge distillation based method~\cite{monti2022many} is suggested to learn trajectory encoding from differenct time step configured trajectory data.
% We apply this method by distillate knowledge from source dataset (WOMD) time configured data to target dataset (nuScenes) configured data.
% The 
Lastly, we apply domain-adpation method which is feature align loss $\mathcal{L}_{align}$ between source and target dataset following~\cite{9880042}, where MMD loss with RBF kernel is used for the distance function.
% Here, for the distance function, MMD loss with RBF kernel is used.
While the original paper tackled unsupervised domain adaptation problem, we provide labels for the target dataset for a fair comparison with other methods.
However, applying $\mathcal{L}_{align}$ hinder prediction loss from target dataset supervision and had adverse effects on the prediction performance.
% Because of this, applying $\mathcal{L}_{align}$ decreases accuracy as it hinders regression loss from target dataset supervision.

\label{result:other_method}
\begin{table}[t]
\centering
\begin{tabular}{cccccc} \toprule
\small{Baseline} & RD    & \small{S $\to$ T} & \small{T $\to$ S} & $\mathcal{L}_{align}$ & SDE   \\ \midrule
0.950    & 0.944 & 0.987                        & 0.947                        & 0.971            & 0.912 \\ \bottomrule
\end{tabular}
\caption{The effectiveness of time-wise continuous representation of NSDE compared to other methods. The digits represent mADE$_{10}$ on nuScenes \textit{val} set for models trained on N+W. Lower is better.}
% \vspace{-10pt}
\label{tab:result_time_aug}
\end{table}

\subsection{Uncertainty handling ability}
\label{result:uncertainty}

% 우리 방법은 uncertain sample 인식 및 걔네에 큰 노이즈를 줌으로서 regularization의 효과를 극대화함으로서 noise에 대한 robustness를 높인다. 따라서, 우리 방법이 잘 동작하는지 확인키 위해 uncertain sample 인식 성능을 분석하였다. 인식 성능은 SDE-net의 방법을 따랐고, 자세한건 supp.

\begin{figure}[t]
\begin{minipage}[c]{0.49\columnwidth}
    \centering
    \includegraphics[width=0.95\columnwidth]{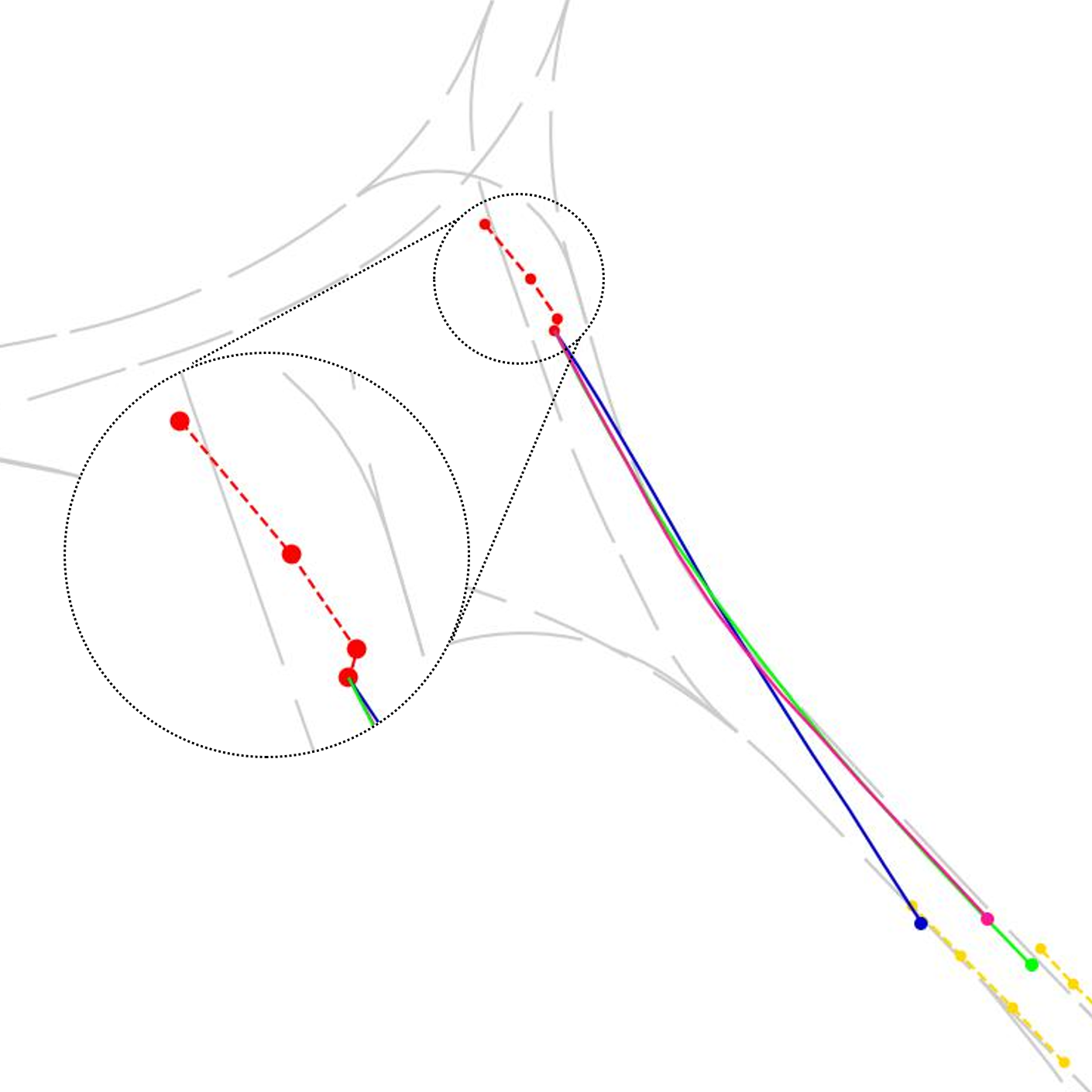}
\end{minipage}%%
    \hfill
\begin{minipage}[c]{0.49\columnwidth}
    \centering
    \includegraphics[width=0.95\columnwidth]{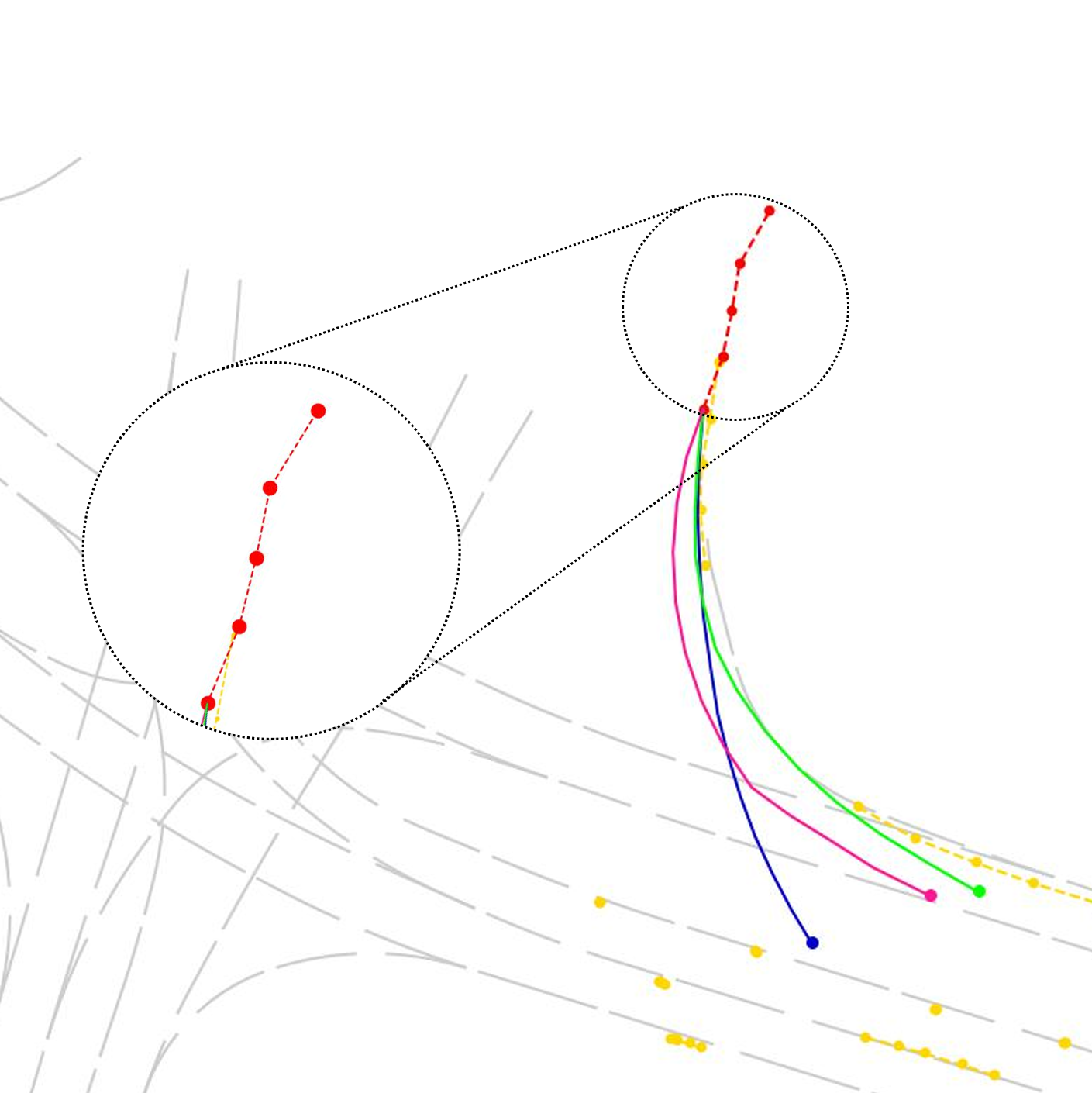}
\end{minipage}
\begin{minipage}[c]{0.49\columnwidth}
    \centering
    \includegraphics[width=0.95\columnwidth]{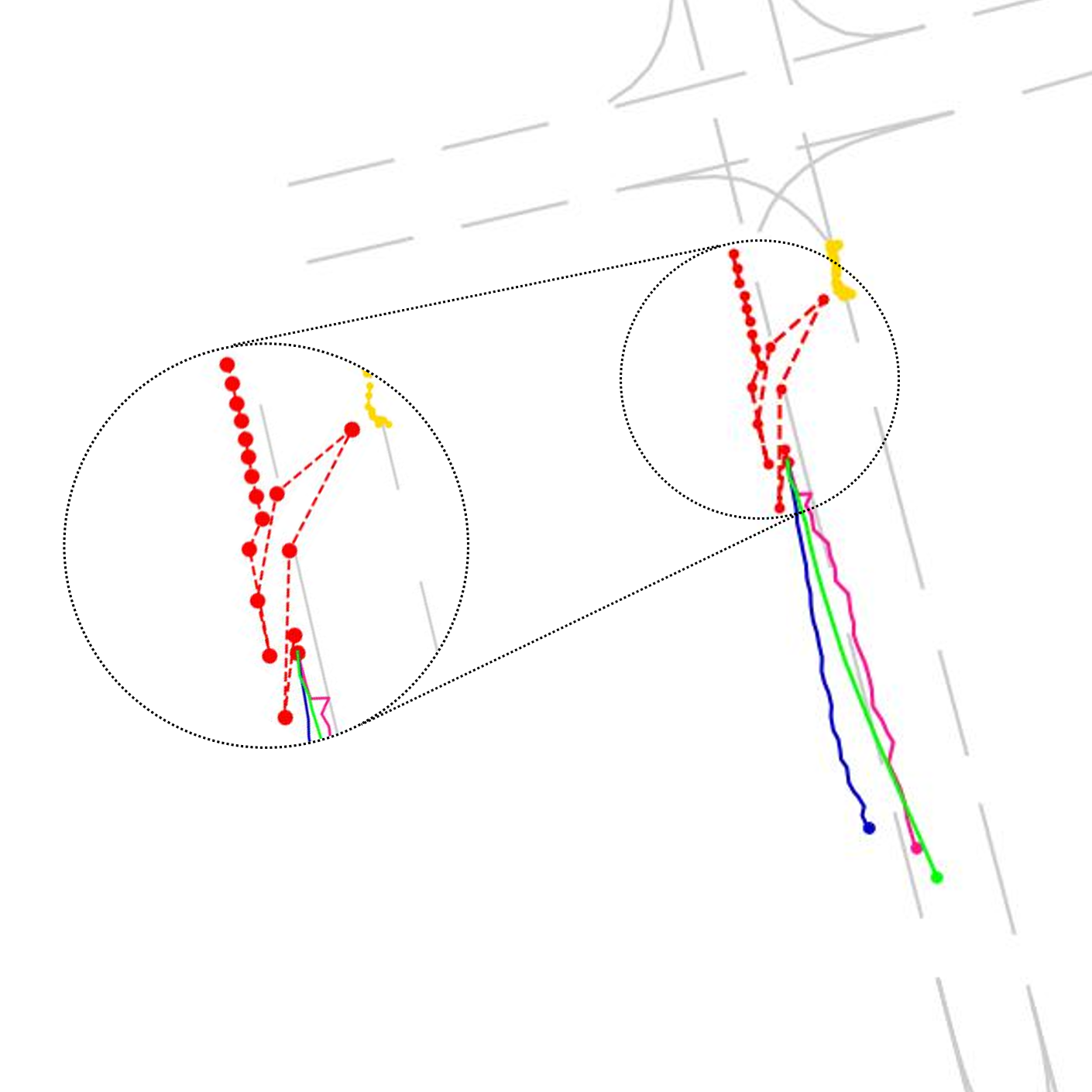}
\end{minipage}%%
\hfill
\begin{minipage}[c]{0.49\columnwidth}
    \centering
    \includegraphics[width=0.95\columnwidth]{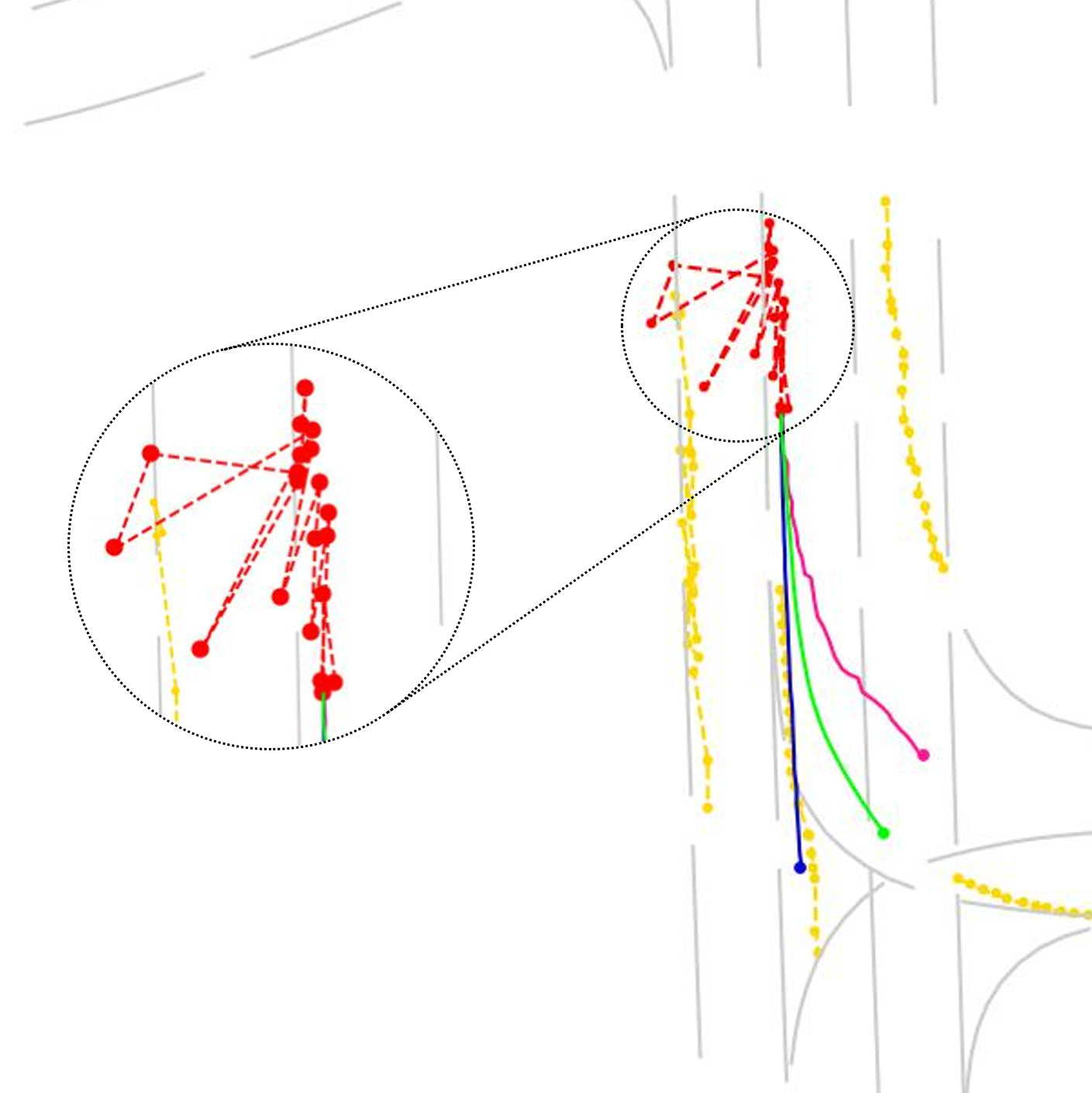}
\end{minipage}
\begin{minipage}[c]{0.49\columnwidth}
    \centering
    \includegraphics[width=0.95\columnwidth]{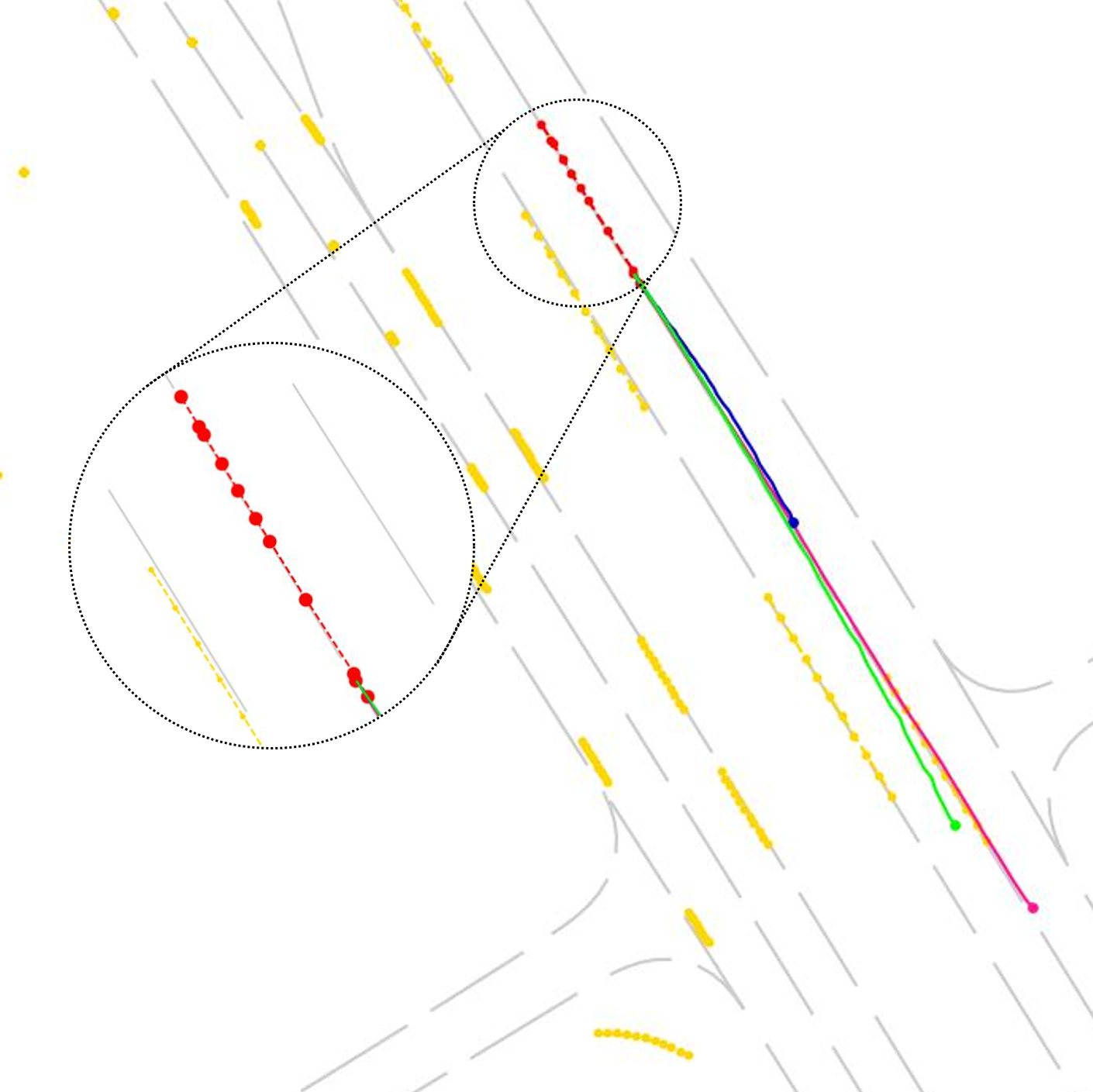}
\end{minipage}%%
\hfill
\begin{minipage}[c]{0.49\columnwidth}
    \centering
    \includegraphics[width=0.95\columnwidth]{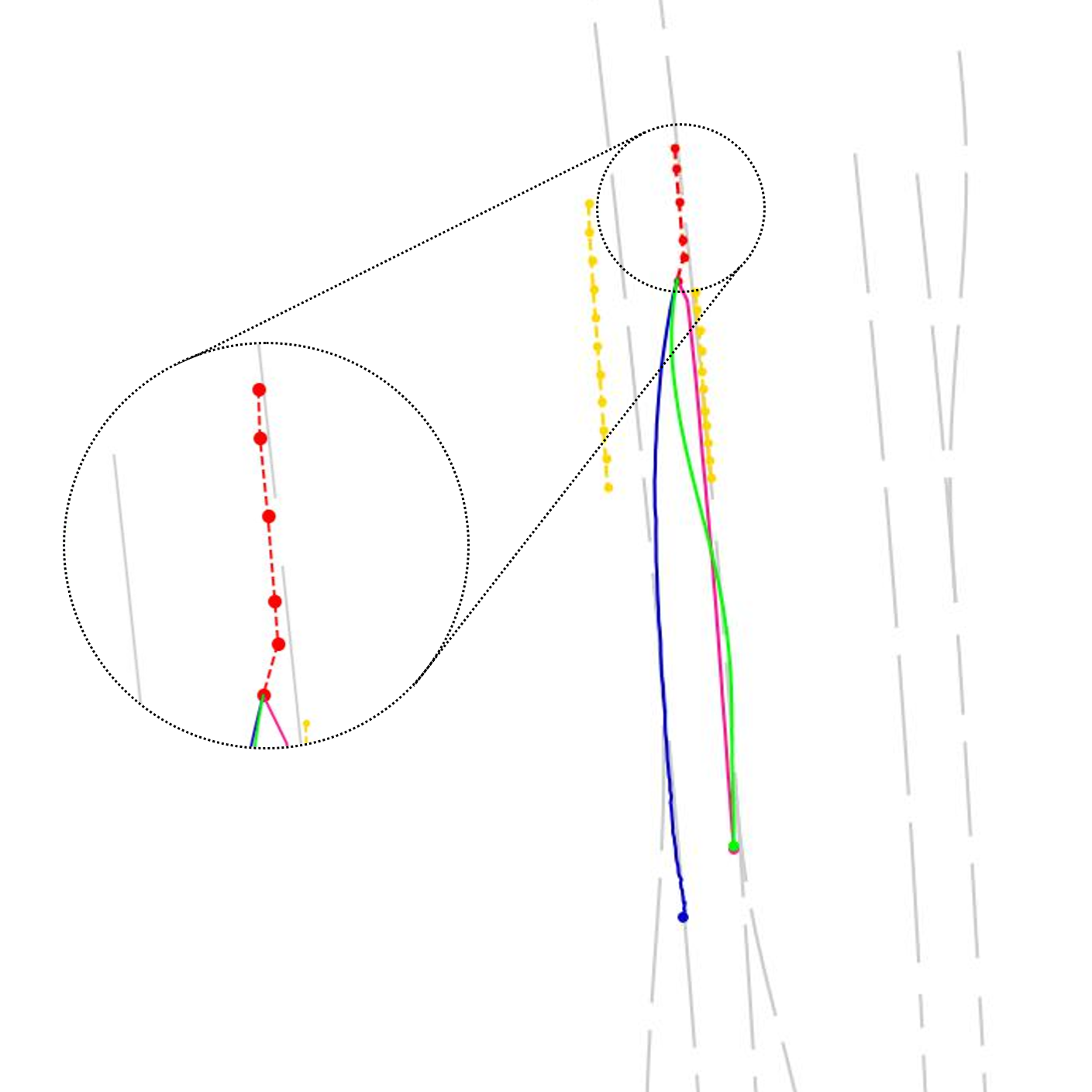}
\end{minipage}
\caption{
Prediction results against uncertain trajectory inputs. 
The dotted line is past trajectories of road actors.
% Among the \textcolor{yellow}{yellow} normal trajectories, the \textcolor{red}{red} ones are actors which are recognized as uncertain samples following SDE-Net framework. 
Among the \textcolor{yellow}{yellow} normal trajectories, the \textcolor{red}{red} ones are samples recognized as uncertain by the SDE-Net framework. 
Their GT future paths are plotted in \textcolor{magenta}{magenta}, prediction from baseline (HiVT) and ours are plotted in \textcolor{blue}{blue} and \textcolor{green}{green}, respectively.
Among 10 predictions of both models, most closest ones to the GT are plotted.
Each row represents nuScenes (1st), Argoverse (2nd), and WOMD (3rd).
}
% \vspace{-10pt}
\label{fig:result_uncertain}
\end{figure}

Our NSDE intensifies SDE's regularization effects by recognizing uncertain samples and assigning them large browninan motion weighting.
% in order to raise robustness against noise. 
Our method relies on the recognition of uncertain samples, therefore we quantify the recognized uncertainty to assess our method's adequate operation.
% For this purpose, we follow SDE-Net's uncertainty quantification method to recognize uncertain sample.
The details of uncertainty quantification process are explained in the supplementary materials.
% Following SDE-Net, we quantify sample uncertainty based on its scale of diffusion network output.
% As explained in the Method section, our NSDE first recognizes uncertain samples based on their scale of diffusion network output, following SDE-Net.
% Details on the process of quantification could be found in the supplementary materials.
% Then, these uncertain samples are assigned larger browninan motion term when solving SDE, thus improving robustness against them.
% Therefore, the recognition of uncertainty is a crucial first step of our method, thus evaluated both qualitatively and quantitatively.
% To check whether our model can recognize the uncertainty due to tracklet noise, we quantify the uncertainty in all datasets with the model trained with our method. 
For a qualitative review, Fig.~\ref{fig:result_uncertain} plots uncertain samples in red lines and others in yellow, thresholded by average standard deviation value of 0.06.
% In Fig.~\ref{fig:result_uncertain}, we plotted uncertain samples in red lines that has average standard deviation value over 0.06, and others in yellow.
In the nuScenes samples (1st row), it shows that our model can properly recognize uncertain samples due to tracking error with sudden position change (left) and meandering motion (right).
Other dataset samples also show competent classification of samples with their dataset-specific uncertainties. 
More samples can be found in supplementary material.

% 다음으로, 이렇게 uncertain sample에 부여된 노이즈가 실제로 robustness를 높여주었는지 확인해보자

Here, we analyze whether the diffusion net-based brownian-motion weighting indeed improve the model's robustness against tracklet error.
% Followed by recognition of uncertain samples by the diffusion network, our NSDE is trained with magnitude of brownian motion term determined by the diffusion network's uncertainty quantification.
% In addition to noisy sample detection,
In doing so, we compare the prediction results between our method (green) and the baseline (blue) on \textit{ood} samples as in Fig.~\ref{fig:result_uncertain}.
% The results between baseline HiVT (blue), ours (green), and ground truth (magenta).
% To assess this ability, we compared prediction accuracy of ours and the baseline HiVT on samples recognized as uncertain samples.
% For a quantitative assessment, we compared prediction accuracy of ours and the baseline HiVT on samples recognized as uncertain samples.
% In Fig.~\ref{fig:result_uncertain}, prediction of the baseline (blue) and ours (green) are plotted, along with the GT future path (magenta).
Among 10 predictions for both models, only the most accurate predictions to GT (magenta) are plotted.
% We plotted the most accurate sample among the prediction with the GT.
Predictions of our method are consistently more accurate compared to the baseline's predictions.
% Looking at the results, baseline predictions are sensitive to tracklet errors, yielding inaccuracies; in contrast, ours achieves better accuracy.
Such improvement is also quantitatively compared in Tab.~\ref{tab:result_uncertain_quan} where mADE$_{10}$ is compared between predictions of baseline and ours on normal samples (\textit{ind}) and uncertain samples (\textit{ood}).
% This also can be seen in the quantitative results in Tab.~\ref{tab:result_uncertain_quan} where mADE$_{10}$ is measured for normal samples (\textit{ind}) and uncertain samples (\textit{ood}) using threshold of 0.01 for \textit{ind} and 0.06 for \textit{ood} for average diffusion net outputs.
% Looking at the prediction accuracy gain of both, ours shows larger accuracy gain in \textit{ood}, which shows that our method is more robust against uncertain samples due to tracklet noise. 
Notably, our method exhibits larger accuracy gains in \textit{ood}, demonstrating superior robustness against uncertain samples due to tracklet noise.
\begin{table}[t]
    \centering
    \begin{tabular}{ccc} \toprule
        \tiny{thres: 0.01/0.06} & \textit{ind}   & \textit{ood}   \\ \midrule
        Baseline         & 0.580 & 1.414 \\
        Ours  & 0.551 & 1.333 \\ \midrule
        gain             & 0.029 & \textbf{0.081} \\ \bottomrule
    \end{tabular}
    \caption{Prediction accuracy (mADE$_{10}$) against normal samples (\textit{ind}) and uncertain samples (\textit{ood}) in nuScenes \textit{val} set. Baseline HiVT and the proposed methods trained on N+A are compared. Lower is better.}
    \label{tab:result_uncertain_quan}
\end{table}

\subsection{Ablation study}
\label{ablation}
Ablation on model architecture appears in the upper part of Tab.~\ref{tab:result_ablation}, with HiVT as baseline.
First, we model the past feature at time 0 ($\textup{h}_0^{past}$) as Gaussian latent following general NDE methods.
After obtaining mean and variance from $\textup{h}_0^{past}$ as in VAE, we sample past feature $F$ times. ($F$ is the number of prediction sample, here, set as 10).
This approach severly worsen the performance since non-probability sampling is much better for multi-modal trajectory prediction~\cite{bae2022non}.
% This approach severly worsened the performance as well known as non-probability sampling is much better for multi-modal trajectory prediction~\cite{bae2022non}.
Second, we adjust the number of layers of drift and diffusion network of encoder NSDE and decoder NSDE.
Their number of layer is originally set as 4, and we reduce them as two.
Comparing the results between encoder and decoder, model capacity decline resulted in larger performance drop for the encoder.
We believe such discrepancy comes from higher complexity of encoder's task, as the encoder needs to translate past features while also considering incoming data on certain timesteps via GRU.
% In our analysis, it is due to encoder need to translate past feature considering incoming data through GRU, so its process become more complex task.

The lower part of Tab.~\ref{tab:result_ablation} shows ablation on uncertainty training.
Our model is comprised of shared drift network along with separate diffusion networks and we ablate each component. 
% First, from our famework that sharing drift network and seperate diffusion network across dataset, we conduct experiments that sepereate drift network (w/o share Drift net), and share diffusion network (w/o seperate Diff net). 
Although we lose multi-source training for temporal encoding when separating drift network, performance drop is relatively small since other components of the model are still shared.
% If we separate drift network, we do not fully enjoy the multi-source training for time series encoding, but it still share other part of models, so its performance drop is relatively small.
In case of sharing diffusion network, however, the performance drop is larger since a single diffusion network is insufficient to handle disparate types of noises.
Indeed, noise in argoverse is more severe compared to nuScenes as shown in the second row in Fig.~\ref{fig:result_uncertain}, so
% Since argoverse data is 10$\times$ of nuScenes data, 
the diffusion network is dominated by argoverse data's distribution and results in wrong Brownian noise injection.
% It is because the noise in Argoverse dataset is too severe as shown in the second row in Fig.~\ref{fig:result_uncertain}, so diffusion network trained on such noisy data does not work well on nuScenes data.
In addition, we remove the uncertainty training objective in Eq.~\ref{eq:final_g_obj}, which has also resulted in a considerable performance drop.
The above results consistently reveal that uncertainty handling plays a crucial role during cross-domain trajectory prediction.

\begin{table}[t]
\centering
\begin{tabular}{lll} \toprule
        & Experiments                               & mADE$_{10}$ \\ \midrule
\multirow{3}{*}{\makecell{Model \\ architecture}}   & Gaussian latent         & 1.225       \\
                                                  & Encoder f/g layers=2     &   0.927     \\ 
                                                  & Decoder f/g layers=2     &   0.918     \\ \midrule
\multirow{3}{*}{\makecell{Uncertainty \\ training}} & w/o share Drift net &  0.934      \\
                                                  & w/o separate Diff net    &  0.944      \\
                                                  & w/o $\mathcal{L}_{\text{uncertain}}$   &  0.940      \\ \midrule
                                                  &  Full model                  & 0.913      \\ \bottomrule
\end{tabular}
\caption{Ablation studies on N+A training.}
% \vspace{-10pt}
\label{tab:result_ablation}
\end{table}

\section{Conclusion}
In this paper, we introduce a novel approach to addressing the challenges posed by discrepancies in trajectory datasets. 
By leveraging continuous and stochastic representations within NSDE, the proposed method tackles two key issues: varying time step configurations and different patterns of detection/tracking noise across datasets. 
The continuous representation effectively handles diverse time intervals, enabling seamless adaptation to different dataset structures, while the stochastic aspect accommodates the inherent uncertainties arising from tracklet errors. 
Through experimentation on nuScenes, Argoverse, Waymo, INTERACTION, and WOMD, our NSDE consistently improved upon both regression and goal prediction-based the state-of-the-art methods.

We not only highlight the importance of dataset-specific considerations in trajectory prediction but also introduce a practical solution that bridges the gap between diverse data sources. 
% The adoption of continuous and stochastic representations within the NSDE framework not only demonstrates the ability to handle temporal disparities but also effectively mitigates the impact of detection and tracking noise. 
These contributions underscore the methodology's potential for advancing the reliability and safety of autonomous mobility systems, offering a promising avenue for further research and development in the field.

\section{Acknowledgments}

This work was partially supported by Institute of Information \& Communications Technology Planning \& Evaluation(IITP) grant funded by the Korea government(MSIT) (No.2014-3-00123, Development of High Performance Visual BigData Discovery Platform for Large-Scale Realtime Data Analysis), and by the National Research Foundation of Korea (NRF) grant funded by the Korea government (MSIT) (NRF2022R1A2B5B03002636).

\bibliography{aaai24}

\begin{thebibliography}{50}
\providecommand{\natexlab}[1]{#1}

\bibitem[{Anumasa and Srijith(2022)}]{anumasa2022latent}
Anumasa, S.; and Srijith, P. 2022.
\newblock Latent time neural ordinary differential equations.
\newblock In \emph{Proceedings of the AAAI Conference on Artificial Intelligence}, volume~36, 6010--6018.

\bibitem[{Bae and Jeon(2023)}]{Bae2023ASO}
Bae, I.; and Jeon, H.-G. 2023.
\newblock A Set of Control Points Conditioned Pedestrian Trajectory Prediction.
\newblock In \emph{AAAI Conference on Artificial Intelligence}.

\bibitem[{Bae, Park, and Jeon(2022)}]{bae2022non}
Bae, I.; Park, J.-H.; and Jeon, H.-G. 2022.
\newblock Non-probability sampling network for stochastic human trajectory prediction.
\newblock In \emph{Proceedings of the IEEE/CVF Conference on Computer Vision and Pattern Recognition}, 6477--6487.

\bibitem[{Caesar et~al.(2020)Caesar, Bankiti, Lang, Vora, Liong, Xu, Krishnan, Pan, Baldan, and Beijbom}]{caesar2020nuscenes}
Caesar, H.; Bankiti, V.; Lang, A.~H.; Vora, S.; Liong, V.~E.; Xu, Q.; Krishnan, A.; Pan, Y.; Baldan, G.; and Beijbom, O. 2020.
\newblock nuscenes: A multimodal dataset for autonomous driving.
\newblock In \emph{Proceedings of the IEEE/CVF conference on computer vision and pattern recognition}, 11621--11631.

\bibitem[{Cao et~al.(2023)Cao, Enouen, Wang, Song, Meng, Niu, and Liu}]{cao2023estimating}
Cao, D.; Enouen, J.; Wang, Y.; Song, X.; Meng, C.; Niu, H.; and Liu, Y. 2023.
\newblock Estimating Treatment Effects from Irregular Time Series Observations with Hidden Confounders.
\newblock \emph{arXiv preprint arXiv:2303.02320}.

\bibitem[{Chang et~al.(2019)Chang, Lambert, Sangkloy, Singh, Bak, Hartnett, Wang, Carr, Lucey, Ramanan et~al.}]{chang2019argoverse}
Chang, M.-F.; Lambert, J.; Sangkloy, P.; Singh, J.; Bak, S.; Hartnett, A.; Wang, D.; Carr, P.; Lucey, S.; Ramanan, D.; et~al. 2019.
\newblock Argoverse: 3d tracking and forecasting with rich maps.
\newblock In \emph{Proceedings of the IEEE/CVF conference on computer vision and pattern recognition}, 8748--8757.

\bibitem[{Chen et~al.(2018)Chen, Rubanova, Bettencourt, and Duvenaud}]{NEURIPS2018_69386f6b}
Chen, R. T.~Q.; Rubanova, Y.; Bettencourt, J.; and Duvenaud, D.~K. 2018.
\newblock Neural Ordinary Differential Equations.
\newblock In Bengio, S.; Wallach, H.; Larochelle, H.; Grauman, K.; Cesa-Bianchi, N.; and Garnett, R., eds., \emph{Advances in Neural Information Processing Systems}, volume~31. Curran Associates, Inc.

\bibitem[{De~Brouwer et~al.(2019)De~Brouwer, Simm, Arany, and Moreau}]{de2019gru}
De~Brouwer, E.; Simm, J.; Arany, A.; and Moreau, Y. 2019.
\newblock GRU-ODE-Bayes: Continuous modeling of sporadically-observed time series.
\newblock \emph{Advances in neural information processing systems}, 32.

\bibitem[{Ettinger et~al.(2021)Ettinger, Cheng, Caine, Liu, Zhao, Pradhan, Chai, Sapp, Qi, Zhou et~al.}]{ettinger2021large}
Ettinger, S.; Cheng, S.; Caine, B.; Liu, C.; Zhao, H.; Pradhan, S.; Chai, Y.; Sapp, B.; Qi, C.~R.; Zhou, Y.; et~al. 2021.
\newblock Large scale interactive motion forecasting for autonomous driving: The waymo open motion dataset.
\newblock In \emph{Proceedings of the IEEE/CVF International Conference on Computer Vision}, 9710--9719.

\bibitem[{Ge, Song, and Huang(2023)}]{Ge2023CausalIF}
Ge, C.; Song, S.; and Huang, G. 2023.
\newblock Causal Intervention for Human Trajectory Prediction with Cross Attention Mechanism.
\newblock In \emph{AAAI Conference on Artificial Intelligence}.

\bibitem[{Gilles et~al.(2022)Gilles, Sabatini, Tsishkou, Stanciulescu, and Moutarde}]{Gilles2022UncertaintyEF}
Gilles, T.; Sabatini, S.; Tsishkou, D.~V.; Stanciulescu, B.; and Moutarde, F. 2022.
\newblock Uncertainty estimation for Cross-dataset performance in Trajectory prediction.
\newblock \emph{ArXiv}, abs/2205.07310.

\bibitem[{{google-research}(2021)}]{torchsde_git}
{google-research}. 2021.
\newblock {torchsde}.
\newblock \url{https://github.com/google-research/torchsde}.

\bibitem[{Hochreiter and Schmidhuber(1997)}]{6795963}
Hochreiter, S.; and Schmidhuber, J. 1997.
\newblock Long Short-Term Memory.
\newblock \emph{Neural Computation}, 9(8): 1735--1780.

\bibitem[{Houston et~al.(2021)Houston, Zuidhof, Bergamini, Ye, Chen, Jain, Omari, Iglovikov, and Ondruska}]{houston2021one}
Houston, J.; Zuidhof, G.; Bergamini, L.; Ye, Y.; Chen, L.; Jain, A.; Omari, S.; Iglovikov, V.; and Ondruska, P. 2021.
\newblock One thousand and one hours: Self-driving motion prediction dataset.
\newblock In \emph{Conference on Robot Learning}, 409--418. PMLR.

\bibitem[{Hu et~al.(2023)Hu, Yang, Chen, Li, Sima, Zhu, Chai, Du, Lin, Wang, Lu, Jia, Liu, Dai, Qiao, and Li}]{hu2023_uniad}
Hu, Y.; Yang, J.; Chen, L.; Li, K.; Sima, C.; Zhu, X.; Chai, S.; Du, S.; Lin, T.; Wang, W.; Lu, L.; Jia, X.; Liu, Q.; Dai, J.; Qiao, Y.; and Li, H. 2023.
\newblock Planning-oriented Autonomous Driving.
\newblock In \emph{Proceedings of the IEEE/CVF Conference on Computer Vision and Pattern Recognition}.

\bibitem[{Ivanovic et~al.(2022)Ivanovic, Lin, Shrivastava, Chakravarty, and Pavone}]{9811776}
Ivanovic, B.; Lin, Y.; Shrivastava, S.; Chakravarty, P.; and Pavone, M. 2022.
\newblock Propagating State Uncertainty Through Trajectory Forecasting.
\newblock In \emph{2022 International Conference on Robotics and Automation (ICRA)}, 2351--2358.

\bibitem[{Kidger et~al.(2020)Kidger, Morrill, Foster, and Lyons}]{kidger2020neural}
Kidger, P.; Morrill, J.; Foster, J.; and Lyons, T. 2020.
\newblock Neural controlled differential equations for irregular time series.
\newblock \emph{Advances in Neural Information Processing Systems}, 33: 6696--6707.

\bibitem[{Kong, Sun, and Zhang(2020)}]{kong2020sde}
Kong, L.; Sun, J.; and Zhang, C. 2020.
\newblock SDE-Net: Equipping Deep Neural Networks with Uncertainty Estimates.
\newblock In \emph{International Conference on Machine Learning}, 5405--5415. PMLR.

\bibitem[{Kothari, Kreiss, and Alahi(2021)}]{kothari2021human}
Kothari, P.; Kreiss, S.; and Alahi, A. 2021.
\newblock Human trajectory forecasting in crowds: A deep learning perspective.
\newblock \emph{IEEE Transactions on Intelligent Transportation Systems}, 23(7): 7386--7400.

\bibitem[{Lee et~al.(2022)Lee, Sohn, Moon, Yoon, Kapadia, and Pavlovic}]{Lee_2022_CVPR}
Lee, M.; Sohn, S.~S.; Moon, S.; Yoon, S.; Kapadia, M.; and Pavlovic, V. 2022.
\newblock MUSE-VAE: Multi-Scale VAE for Environment-Aware Long Term Trajectory Prediction.
\newblock In \emph{Proceedings of the IEEE/CVF Conference on Computer Vision and Pattern Recognition (CVPR)}, 2221--2230.

\bibitem[{Li et~al.(2021)Li, Yao, Wenliang, He, Xiao, Yan, Wipf, and Zhang}]{li2021grin}
Li, L.; Yao, J.; Wenliang, L.~K.; He, T.; Xiao, T.; Yan, J.; Wipf, D.; and Zhang, Z. 2021.
\newblock {GRIN}: Generative Relation and Intention Network for Multi-agent Trajectory Prediction.
\newblock In Beygelzimer, A.; Dauphin, Y.; Liang, P.; and Vaughan, J.~W., eds., \emph{Advances in Neural Information Processing Systems}.

\bibitem[{Li et~al.(2020)Li, Wong, Chen, and Duvenaud}]{li2020scalable}
Li, X.; Wong, T.-K.~L.; Chen, R.~T.; and Duvenaud, D. 2020.
\newblock Scalable gradients for stochastic differential equations.
\newblock In \emph{International Conference on Artificial Intelligence and Statistics}, 3870--3882. PMLR.

\bibitem[{Liang et~al.(2021)Liang, Li, Li, Tang, Zhou, and Zou}]{liang2021temporal}
Liang, R.; Li, Y.; Li, X.; Tang, Y.; Zhou, J.; and Zou, W. 2021.
\newblock Temporal pyramid network for pedestrian trajectory prediction with multi-supervision.
\newblock In \emph{Proceedings of the AAAI conference on artificial intelligence}, volume~35, 2029--2037.

\bibitem[{Liu et~al.(2019)Liu, Xiao, Si, Cao, Kumar, and Hsieh}]{liu2019neural}
Liu, X.; Xiao, T.; Si, S.; Cao, Q.; Kumar, S.; and Hsieh, C.-J. 2019.
\newblock Neural sde: Stabilizing neural ode networks with stochastic noise.
\newblock \emph{arXiv preprint arXiv:1906.02355}.

\bibitem[{Malinin et~al.(2021)Malinin, Band, Gal, Gales, Ganshin, Chesnokov, Noskov, Ploskonosov, Prokhorenkova, Provilkov, Raina, Raina, Roginskiy, Shmatova, Tigas, and Yangel}]{malinin2021shifts}
Malinin, A.; Band, N.; Gal, Y.; Gales, M.; Ganshin, A.; Chesnokov, G.; Noskov, A.; Ploskonosov, A.; Prokhorenkova, L.; Provilkov, I.; Raina, V.; Raina, V.; Roginskiy, D.; Shmatova, M.; Tigas, P.; and Yangel, B. 2021.
\newblock Shifts: A Dataset of Real Distributional Shift Across Multiple Large-Scale Tasks.
\newblock In \emph{Thirty-fifth Conference on Neural Information Processing Systems Datasets and Benchmarks Track (Round 2)}.

\bibitem[{Meng et~al.(2022)Meng, Wu, Chen, Cai, Zhou, Yang, and Shen}]{meng2022forecasting}
Meng, M.; Wu, Z.; Chen, T.; Cai, X.; Zhou, X.~S.; Yang, F.; and Shen, D. 2022.
\newblock Forecasting Human Trajectory from Scene History.
\newblock In Oh, A.~H.; Agarwal, A.; Belgrave, D.; and Cho, K., eds., \emph{Advances in Neural Information Processing Systems}.

\bibitem[{Norcliffe et~al.(2020)Norcliffe, Bodnar, Day, Moss, and Li{\`o}}]{norcliffe2020neural}
Norcliffe, A.; Bodnar, C.; Day, B.; Moss, J.; and Li{\`o}, P. 2020.
\newblock Neural ODE Processes.
\newblock In \emph{International Conference on Learning Representations}.

\bibitem[{Park and Park(2020)}]{park2020identifying}
Park, D.; and Park, Y.-H. 2020.
\newblock Identifying Reflected Images From Object Detector in Indoor Environment Utilizing Depth Information.
\newblock \emph{IEEE Robotics and Automation Letters}, 6(2): 635--642.

\bibitem[{Park et~al.(2022)Park, Ryu, Yang, Cho, Kim, and Yoon}]{park2022leveraging}
Park, D.; Ryu, H.; Yang, Y.; Cho, J.; Kim, J.; and Yoon, K.-J. 2022.
\newblock Leveraging {Future} {Relationship} {Reasoning} for {Vehicle} {Trajectory} {Prediction}.
\newblock In \emph{The {Eleventh} {International} {Conference} on {Learning} {Representations}}.

\bibitem[{Park et~al.(2021)Park, Kim, Lee, Choo, Lee, Kim, and Choi}]{park2021vid}
Park, S.; Kim, K.; Lee, J.; Choo, J.; Lee, J.; Kim, S.; and Choi, E. 2021.
\newblock Vid-ode: Continuous-time video generation with neural ordinary differential equation.
\newblock In \emph{Proceedings of the AAAI Conference on Artificial Intelligence}, volume~35, 2412--2422.

\bibitem[{Qian, Kacprzyk, and van~der Schaar(2022)}]{qian2022dcode}
Qian, Z.; Kacprzyk, K.; and van~der Schaar, M. 2022.
\newblock D-{CODE}: Discovering Closed-form {ODE}s from Observed Trajectories.
\newblock In \emph{International Conference on Learning Representations}.

\bibitem[{Rubanova, Chen, and Duvenaud(2019)}]{rubanova2019latent}
Rubanova, Y.; Chen, R.~T.; and Duvenaud, D.~K. 2019.
\newblock Latent ordinary differential equations for irregularly-sampled time series.
\newblock \emph{Advances in neural information processing systems}, 32.

\bibitem[{Saleh et~al.(2021)Saleh, Aliakbarian, Rezatofighi, Salzmann, and Gould}]{saleh2021probabilistic}
Saleh, F.; Aliakbarian, S.; Rezatofighi, H.; Salzmann, M.; and Gould, S. 2021.
\newblock Probabilistic tracklet scoring and inpainting for multiple object tracking.
\newblock In \emph{Proceedings of the IEEE/CVF Conference on Computer Vision and Pattern Recognition}, 14329--14339.

\bibitem[{Salzmann et~al.(2020)Salzmann, Ivanovic, Chakravarty, and Pavone}]{salzmann2020trajectron++}
Salzmann, T.; Ivanovic, B.; Chakravarty, P.; and Pavone, M. 2020.
\newblock Trajectron++: Dynamically-feasible trajectory forecasting with heterogeneous data.
\newblock In \emph{Computer Vision--ECCV 2020: 16th European Conference, Glasgow, UK, August 23--28, 2020, Proceedings, Part XVIII 16}, 683--700. Springer.

\bibitem[{Shi et~al.(2022)Shi, Wang, Long, Zhou, Zheng, Zheng, and Hua}]{shi2022social}
Shi, L.; Wang, L.; Long, C.; Zhou, S.; Zheng, F.; Zheng, N.; and Hua, G. 2022.
\newblock Social interpretable tree for pedestrian trajectory prediction.
\newblock In \emph{Proceedings of the AAAI Conference on Artificial Intelligence}, volume~36, 2235--2243.

\bibitem[{Tang et~al.(2021)Tang, Zhong, Neumann, Wang, Chen, and Zhang}]{tang2021collaborative}
Tang, B.; Zhong, Y.; Neumann, U.; Wang, G.; Chen, S.; and Zhang, Y. 2021.
\newblock Collaborative Uncertainty in Multi-Agent Trajectory Forecasting.
\newblock In Beygelzimer, A.; Dauphin, Y.; Liang, P.; and Vaughan, J.~W., eds., \emph{Advances in Neural Information Processing Systems}.

\bibitem[{Tzen and Raginsky(2019)}]{tzen2019neural}
Tzen, B.; and Raginsky, M. 2019.
\newblock Neural stochastic differential equations: Deep latent gaussian models in the diffusion limit.
\newblock \emph{arXiv preprint arXiv:1905.09883}.

\bibitem[{Vaswani et~al.(2017)Vaswani, Shazeer, Parmar, Uszkoreit, Jones, Gomez, Kaiser, and Polosukhin}]{vaswani2017attention}
Vaswani, A.; Shazeer, N.; Parmar, N.; Uszkoreit, J.; Jones, L.; Gomez, A.~N.; Kaiser, {\L}.; and Polosukhin, I. 2017.
\newblock Attention is all you need.
\newblock \emph{Advances in neural information processing systems}, 30.

\bibitem[{Wang et~al.(2022{\natexlab{a}})Wang, Chen, Wang, and Wang}]{Wang_2022_CVPR}
Wang, C.; Chen, X.; Wang, J.; and Wang, H. 2022{\natexlab{a}}.
\newblock ATPFL: Automatic Trajectory Prediction Model Design Under Federated Learning Framework.
\newblock In \emph{Proceedings of the IEEE/CVF Conference on Computer Vision and Pattern Recognition (CVPR)}, 6563--6572.

\bibitem[{Wang et~al.(2022{\natexlab{b}})Wang, Hu, Sun, Zhan, Tomizuka, and Liu}]{wang2022transferable}
Wang, L.; Hu, Y.; Sun, L.; Zhan, W.; Tomizuka, M.; and Liu, C. 2022{\natexlab{b}}.
\newblock Transferable and adaptable driving behavior prediction.
\newblock \emph{arXiv preprint arXiv:2202.05140}.

\bibitem[{Wang et~al.(2023)Wang, Wang, Yan, and Wang}]{Wang2023WSiPWS}
Wang, R.; Wang, S.; Yan, H.; and Wang, X. 2023.
\newblock WSiP: Wave Superposition Inspired Pooling for Dynamic Interactions-Aware Trajectory Prediction.
\newblock In \emph{AAAI Conference on Artificial Intelligence}.

\bibitem[{Wen, Wang, and Metaxas(2022)}]{10.1007/978-3-031-20047-2_13}
Wen, S.; Wang, H.; and Metaxas, D. 2022.
\newblock Social ODE: Multi-agent Trajectory Forecasting withÂ Neural Ordinary Differential Equations.
\newblock In Avidan, S.; Brostow, G.; Ciss{\'e}, M.; Farinella, G.~M.; and Hassner, T., eds., \emph{Computer Vision -- ECCV 2022}, 217--233. Cham: Springer Nature Switzerland.
\newblock ISBN 978-3-031-20047-2.

\bibitem[{Weng et~al.(2022)Weng, Ivanovic, Kitani, and Pavone}]{9879091}
Weng, X.; Ivanovic, B.; Kitani, K.; and Pavone, M. 2022.
\newblock Whose Track Is It Anyway? Improving Robustness to Tracking Errors with Affinity-based Trajectory Prediction.
\newblock In \emph{2022 IEEE/CVF Conference on Computer Vision and Pattern Recognition (CVPR)}, 6563--6572.

\bibitem[{Weng, Ivanovic, and Pavone(2022)}]{Weng2022_MTP}
Weng, X.; Ivanovic, B.; and Pavone, M. 2022.
\newblock {MTP: Multi-hypothesis Tracking and Prediction for Reduced Error Propagation}.
\newblock \emph{IV}.

\bibitem[{Westny et~al.(2023)Westny, Oskarsson, Olofsson, and Frisk}]{10143287}
Westny, T.; Oskarsson, J.; Olofsson, B.; and Frisk, E. 2023.
\newblock MTP-GO: Graph-Based Probabilistic Multi-Agent Trajectory Prediction with Neural ODEs.
\newblock \emph{IEEE Transactions on Intelligent Vehicles}, 1--14.

\bibitem[{Wu et~al.(2023)Wu, Wang, Zhou, Duan, Hua, and Tang}]{Wu2023MultiStreamRL}
Wu, Y.; Wang, L.; Zhou, S.; Duan, J.; Hua, G.; and Tang, W. 2023.
\newblock Multi-Stream Representation Learning for Pedestrian Trajectory Prediction.
\newblock In \emph{AAAI Conference on Artificial Intelligence}.

\bibitem[{Xu et~al.(2022)Xu, Wang, Wang, and Fu}]{9880042}
Xu, Y.; Wang, L.; Wang, Y.; and Fu, Y. 2022.
\newblock Adaptive Trajectory Prediction via Transferable GNN.
\newblock In \emph{2022 IEEE/CVF Conference on Computer Vision and Pattern Recognition (CVPR)}, 6510--6521.

\bibitem[{Ye, Zhou, and Wang(2023)}]{Ye2023ImprovingTG}
Ye, L.; Zhou, Z.; and Wang, J. 2023.
\newblock Improving the Generalizability of Trajectory Prediction Models with Fren{\'e}t-Based Domain Normalization.
\newblock \emph{2023 IEEE International Conference on Robotics and Automation (ICRA)}, 11562--11568.

\bibitem[{Zhan et~al.(2019)Zhan, Sun, Wang, Shi, Clausse, Naumann, K\"ummerle, K\"onigshof, Stiller, de~La~Fortelle, and Tomizuka}]{interactiondataset}
Zhan, W.; Sun, L.; Wang, D.; Shi, H.; Clausse, A.; Naumann, M.; K\"ummerle, J.; K\"onigshof, H.; Stiller, C.; de~La~Fortelle, A.; and Tomizuka, M. 2019.
\newblock {INTERACTION} {Dataset}: {An} {INTERnational}, {Adversarial} and {Cooperative} {moTION} {Dataset} in {Interactive} {Driving} {Scenarios} with {Semantic} {Maps}.
\newblock \emph{arXiv:1910.03088 [cs, eess]}.

\bibitem[{Zhou et~al.(2022)Zhou, Ye, Wang, Wu, and Lu}]{Zhou_2022_CVPR}
Zhou, Z.; Ye, L.; Wang, J.; Wu, K.; and Lu, K. 2022.
\newblock HiVT: Hierarchical Vector Transformer for Multi-Agent Motion Prediction.
\newblock In \emph{Proceedings of the IEEE/CVF Conference on Computer Vision and Pattern Recognition (CVPR)}, 8823--8833.

\end{thebibliography}

\end{document}